\documentclass{article}
\usepackage[a4paper,left=1in,top=1in,right=1in,bottom=1in,nohead]{geometry}
\usepackage[english]{babel}
\usepackage{lmodern}
\usepackage[T1]{fontenc}  
\usepackage[utf8]{inputenc}
\usepackage{amsmath}
\usepackage{amssymb}
\usepackage{amsfonts}
\usepackage{mathtools}
\usepackage[mathlines, modulo]{lineno}
\usepackage{graphicx}
\usepackage{url}
\usepackage[export]{adjustbox}
\usepackage{xcolor}
\usepackage{booktabs}
\usepackage{natbib}
\usepackage{nameref}

\DeclareMathOperator*{\argmin}{argmin}

\title{Categorical Perception: \\A Groundwork for Deep Learning}
\author{Laurent Bonnasse-Gahot$^{1,*}$ and Jean-Pierre Nadal$^{1,2}$}
\date{
\normalsize
	(1) Centre d'Analyse et de Math\'ematique Sociales \\
	CAMS, UMR 8557 CNRS-EHESS \\
	\'Ecole des Hautes \'Etudes en Sciences Sociales \\
	54 bd. Raspail, 75006 Paris, France \\
	(2) Laboratoire de Physique de l'ENS \\
	LPENS, UMR 8023, CNRS - ENS Paris - PSL University - SU - Universit\'e de Paris\\
	\'Ecole Normale Sup\'erieure \\
	24 rue Lhomond,  75005 Paris, France \\
	$\,$\\
	(*) Corresponding author (\texttt{lbg@ehess.fr})\\
	$\,$\\
	Dated: November 15th, 2021\\
This is an author-produced version of an article accepted for publication in Neural Computation \url{https://doi.org/10.1162/neco_a_01454}
}

\begin{document}
\maketitle
\thispagestyle{empty}

\noindent\rule{\textwidth}{0.7pt}
\begin{abstract}
Classification is one of the major tasks that deep learning is successfully tackling. Categorization is also a fundamental cognitive ability. A well-known perceptual consequence of categorization in humans and other animals, called categorical perception, is notably characterized by a within-category compression and a between-category separation: two items, close in input space, are perceived closer if they belong to the same category than if they belong to different categories. Elaborating on experimental and theoretical results in cognitive science, here we study categorical effects in artificial neural networks. We combine a theoretical analysis that makes use of mutual and Fisher information quantities, and a series of numerical simulations on networks of increasing complexity. These formal and numerical analyses provide insights into the geometry of the neural representation in deep layers, with expansion of space near category boundaries and contraction far from category boundaries. We investigate categorical representation by using two complementary approaches: one mimics experiments in psychophysics and cognitive neuroscience by means of morphed continua between stimuli of different categories, while the other introduces a categoricality index that, for each layer in the network, quantifies the separability of the categories at the neural population level. We show on both shallow and deep neural networks that category learning automatically induces categorical perception. We further show that the deeper a layer, the stronger the categorical effects. As an outcome of our study, we propose a coherent view of the efficacy of different heuristic practices of the dropout regularization technique. More generally, our view, which finds echoes in the neuroscience literature, insists on the differential impact of noise in any given layer depending on the geometry of the neural representation that is being learned, \textit{i.e.} on how this geometry reflects the structure of the categories.\\

\noindent \textit{Keywords:} deep learning; categorical perception; neuronal noise; Fisher information; neural geometry; dropout
\end{abstract}
\noindent\rule{\textwidth}{0.8pt}

\section{Introduction} 
One of the main tasks tackled with deep learning methods is the one of categorization: classification of images as representing specific objects, identification of words for speech recognition, etc. \citep[see][for reviews]{lecun2015deep, schmidhuber2015deep}. The identification of categories is a central topic in cognitive science. In his book ``Categorical Perception: The Groundwork of Cognition'', \citet{Harnad_1987} shows how central and fundamental to cognition categorization is. A well-studied perceptual consequence of categorization in humans and other animals is characterized by a sharp transition in the identification responses and by greater cross-category than within-category discrimination, a phenomenon called categorical perception~\citep[see also][for a review]{Repp_1984}. It originates in the field of speech perception: the seminal work of \citet{Liberman_etal_1957} demonstrated that American English-speaking subjects are better at discriminating between a /ba/ and a /da/ than between two different /ba/ tokens, even when the magnitude of the physical difference between the two stimuli is equal. Authors have subsequently shown that such effects are not specific to English, occurring in any language, but with respect to its own structure, as distinct phonemic systems entail different discrimination abilities \citep{abramson1970discriminability, goto1971auditory}. Categorical perception was also found to be not specific to speech \citep{cross1965identification,burns1978categorical,Bornstein_Korda_1984,Goldstone_1994,Beale_Keil_1995}, nor even to humans \citep{nelson1989categorical, Kluender_etal_1998, caves2018categorical}.\\

Previous computational work has shown that categorical perception also happens in artificial neural networks \citep{anderson1977distinctive, padgett1998simple, tijsseling1997warping, Damper_Harnad_2000}, with internal representations organized into more or less well-separated clusters \citep{mcclelland2003parallel, tong2008fusiform,olah2015visualizing}. A noisy neural classifier, either biological or artificial, that aims at minimizing the probability of misclassifying an incoming stimulus, has to deal with two sources of uncertainty. One is due to the intrinsic overlap between categories (in the relevant case where classifying these stimuli is not an obvious task). The other one stems from the variability of the response of the neurons to a stimulus. Intuitively, one might want to reduce the neural uncertainty in the regions of the input space where the overlap between categories is already a major source of confusion, \textit{ie} the regions of boundaries between classes\footnote{Throughout this paper, we will interchangeably use the two terms `stimulus' and `input', and, likewise, the two terms `categories' and `classes'.}. In \citet{LBG_JPN_2008}, taking an information theoretic approach, in the context of a biologically motivated neural architecture with a large number of coding cells, we quantitatively showed how these two quantities interact precisely, and how as a consequence category learning induces an expansion of the representation space between categories, \textit{i.e.}, categorical perception. Reducing neuronal noise in the region where the chance of misclassifying a new stimulus is highest, by mistakenly crossing the decision boundary, thus lowers the probability of error. Based on our previous studies of categorical perception \citep{LBG_JPN_2008,LBG_JPN_2012}, here we introduce a framework for the analysis of categorization in deep networks. We analyze how categorization builds up in the course of learning and across layers. This analysis reveals the geometry of neural representations in deep layers after learning. We also show that categorical perception is a gradient phenomenon as a function of depth: the deeper, the more pronounced the effect -- from the first hidden layer that might show no specific effect of categorical perception to the last, decisional, layer that exhibits full categorical behavior.\\

So far we mentioned noise as a nuisance that we have to deal with, following the common signal processing tradition. Conversely, a great deal of work has shown that neuronal noise, whether small or large depending on the context, can be desirable, helping a system to learn more efficiently, more robustly, with a better generalization ability. In the field of artificial neural networks, many studies in the 1990s have shown that input noise helps a network to overcome overfitting \citep[see e.g.][]{holmstrom1992using, matsuoka1992noise, bishop1995training, an1996effects}. Authors have also shown that noise can have the effect of revealing the structure of the data, as for learning a rule from examples \citep{seung1992statistical} or in independent component analysis \citep{nadal1994nonlinear}, a fact recently put forward by \citet{schwartz-ziv2017opening} within the framework of the information bottleneck approach. One important ingredient in the \citet{krizhevsky2012imagenet} paper that ignited the recent revival of connectionism in the past decade is the use of multiplicative Bernoulli noise in the hidden activations, a technique called dropout \citep{srivastava2014dropout}. Our study of categorical perception in artificial networks leads us to suggest that a key ingredient making dropout beneficial is that the effect of this noise is not uniform with respect to the input space, but depends on the structure and relation between categories. Importantly, our findings allow us to more generally understand the effect of noise in any given layer as being dependent on the current neural representation, hence as not having the same benefit depending on layer type, layer depth and on the course of learning (initial vs. learned stage).\\

For our analysis of artificial neural networks, we consider two ways of extracting and visualizing the categorical nature of the encoding resulting from learning. The first one consists in mimicking a standard experimental protocol used for the study of categorical perception in humans and other animals. We generate smooth continua interpolating from one category to another, and we look at the structure of the neural activity as one moves along a continuum. We show how, following learning, the representation space is enlarged near the boundary between categories, thus revealing categorical perception: distance between the neural representations of two items is greater when these items are close to the boundary between the classes, compared to the case where they are drawn from a same category. We also show that these categorical effects are more marked the deeper the layer of the network. The second one consists in measuring a categoricality index which quantifies, for each layer, how much the neural representation as a whole is specific to the coding of categories. We show that categoricality gets greater following learning, and increases with the depth of the layer, paralleling the results found by means of the morphed continua. We also observe that convolutional layers, while containing categorical information, are less categorical than their dense counterparts.\\

The paper is organized as follows. In Section~\ref{sec:motivations}, we review empirical studies on categorical perception, and summarize theoretical results that we obtained on the neural basis of this phenomenon. We explain how these findings motivate the present study of categorization in artificial neural networks. In Section~\ref{sec:results}, we conduct computational experiments that confirm the relevance of the study of categorical perception in the understanding of artificial neural networks. We consider several examples gradually going up in task complexity and network sizes. We first illustrate the emergence of categorical perception in artificial neural networks on one- and two- dimensional toy examples. Moving to the MNIST dataset \citep{lecun1998gradient}, a commonly used large database of handwritten digits, we show in Section~\ref{sec:mnist} how categorical perception emerges in an artificial neural network that learns to classify handwritten digits. In Section~\ref{sec:naturalimage} we extend our analysis to very deep networks trained on natural images, working with the Kaggle Dogs vs. Cats dataset, and with the ImageNet dataset \citep{deng2009imagenet,ILSVRC15}. In Section~\ref{sec:dropout}, we discuss many heuristics practices in the use of dropout, along the lines mentioned above. Finally, in Section~\ref{sec:neuro}, we also suggest that our results might in turn shed light on the psychological and neuroscientific study of categorical perception. We provide technical details in the Materials and Methods section and supplementary information in Appendices. 

\section{Insights from cognitive science} 
\label{sec:motivations}
\subsection{Experimental results on categorical perception}
\label{sec:CP}
In the psychological and cognitive science literature, a standard way to look at categorical perception is to present different stimuli along a continuum that evenly interpolate between two stimuli drawn from two different categories. Let us consider a few illustrative examples. In their seminal study on speech perception, \citet{Liberman_etal_1957} generated a /ba/--/da/ synthetic speech continuum by evenly varying the second formant transition, which modulates the perception of the place of articulation. Studying categorical perception of music intervals, \citet{burns1978categorical} considered a continuum that interpolates from a minor third to a major third to a perfect fourth. Turning to vision, \citet{Bornstein_Korda_1984} considered a blue to green continuum, while \citet{Goldstone_1994} made use of rectangles that vary either in brightness or in size. In \citet{Beale_Keil_1995}, the authors generated a continuum of morphed faces interpolating between individual exemplars of familiar faces, from Kennedy to Clinton for instance. As a final example, we cite the monkey study by \citet{freedman2001categorical} that makes use of a dog to cat morphed continuum. In these sorts of studies, experimentalists typically measure category membership for all items along the considered continuum, discrimination between neighboring stimuli, as well as reaction times during a categorization task. Of interest for the present paper are the behaviours in  identification and discrimination tasks observed in these experiments. Although the physical differences in the stimuli change in a continuous way, the identification changes abruptly in a narrow domain near the category boundary while discrimination is better near the category boundary than well inside a category, which is the hallmark of categorical perception.\\

\subsection{Optimal coding entails categorical perception}
\label{sec:CP_mod}
If the neural representations have been optimized for the identification of categories, as e.g. in speech perception, these behavioral performances are intuitive. The goal being to decide to which category the stimulus belongs to, far from a category boundary in stimulus space, there is no need to have a precise identification of the stimulus itself: two nearby stimuli may not be discriminated. On the contrary, in the vicinity of a boundary, the likelihood of a category is strongly affected by any small shift in stimulus space. Thus, one expects the neural code, if optimized in view of the categorization task, to provide a finer representation of the stimuli near a class boundary than within a category.\\

In previous work~\citep{LBG_JPN_2008,LBG_JPN_2012}, we formalized and made quantitative these arguments through modeling of the neural processing (see \ref{sec:appendix_model} for a summary). In empirical studies, authors find stimulus encoding layers where the categorical information is distributed among many cells, no single cell carrying information that fully characterizes a category. On the contrary, in what seems to be the decisional layer, single cells are specific to a single category~\citep[see e.g.][]{kreiman2000category, freedman2001categorical, meyers2008dynamic}. Based on this  neuroscience data, we considered a simplified architecture: a coding layer with a distributed representation, and a decision layer. The coding layer represents the last encoding stage before the decision can be taken, and may correspond to a high level in the neural processing. Hence the input to this layer is not the actual stimulus but some projection of it on some relevant dimensions. Taking a Bayesian and information theoretic approach, we showed that efficient coding (targeting optimal performance in classification) leads to categorical perception (in particular better discrimination near the class boundaries). More precisely, we showed that in order to optimize the neural representation in view of identifying the category, one should maximize the mutual information between the categories and the neural code. This in turn, in the limit of a large number of coding cells, implies that one has to minimize a coding cost which, in qualitative terms, can be written: 
\begin{equation}
\mathcal{\overline{C}}_{\text{coding}} = \frac{1}{2} \left <\frac{\text{\small categorization uncertainty}}{\text{\small neural sensitivity}} \right>
\end{equation}
where the brackets $<.>$ denote the average over the space of relevant dimensions $x$. In formal terms, 
\begin{equation}
\mathcal{\overline{C}}_{\text{coding}} = \frac{1}{2} \int  \frac{F_{\text{cat}}(x)}{F_{\text{code}}(x)} \;p(x)\,dx
\label{eq:midiff}
\end{equation}
where $F_{\text{code}}(x)$ and $F_{\text{cat}}(x)$ are Fisher information quantities. The quantity $F_{\text{code}}(x)$ represents the sensitivity of the neural code to a small change in stimulus $x$. Larger Fisher information means greater sensitivity. The quantity $F_{\text{cat}}(x)$ represents the categorical sensitivity, that is, how much the probability that the stimulus belongs to a given category changes for small variation in $x$ values. Each Fisher information quantity defines a metric over the space $x$ (more exactly, over spaces of probabilities indexed by $x$). Along a path in stimulus space, $F_{\text{cat}}$ quantifies the change in categorical specificity of $x$, and $F_{\text{code}}$ how much the neural activity changes locally. Efficient coding with respect to optimal classification is thus obtained by essentially matching the two metrics. Since $F_{\text{cat}}$ is larger near a class boundary, this should also be the case for $F_{\text{code}}(x)$.\\

The Fisher information  $F_{\text{code}}(x)$ is directly related to the discriminability $d'$ measured in psychophysical experiments \cite[see e.g.,][]{green1966signal}. Within the framework of Signal Detection Theory, $d'$  characterizes the ability to discriminate between two stimuli $x_1=x$ and $x_2=x+\delta x$. If one assumes optimal Bayesian decoding, then this behavioural quantity $d'$ is equal to $|\delta x| \sqrt{F_{\text{code}}(x)}$ \citep{seung1993simple}. Thus, category learning entails greater Fisher information  $F_{\text{code}}(x)$ between categories, hence better cross-category than within-category discrimination, leading to the so-called \textit{categorical perception}. 

\subsection{Category learning and local neural geometry}
\label{sec:geometry}
One may define the dissimilarity $D_{\text{neural}}(x_1, x_2)$ between the stimuli $x_1$ and $x_2$ at the neural level, that is in the space defined by the neural activations, by a distance or a dissimilarity measure between the distributions of activities that they evoke, $D_{\text{neural}}(x_1, x_2) \equiv d(P(\mathbf{r}|x_1)||P(\mathbf{r}|x_2))$. Natural choices for $d(.||.)$ are f-divergences, among which the Kullback-Leibler divergence $D_{KL}(P(\mathbf{r}|x_1)||P(\mathbf{r}|x_2))$, or the symmetrised Kullback-Leibler divergence, $D_{KL}(P(\mathbf{r}|x_1)||P(\mathbf{r}|x_2)) + D_{KL}(P(\mathbf{r}|x_2)||P(\mathbf{r}|x_1))$. In such cases, for small $\delta x = x_2 - x_1$, the distance is proportional to the Fisher information. For, e.g., the symmetrised Kullback-Leibler divergence,
\begin{equation}
D(x, x+\delta x) = (\delta x)^2\, F_{\text{code}}(x).
\end{equation}
Thus, the Fisher information $F_{\text{code}}(x)$  provides the local neural geometry: a larger $F_{\text{code}}(x)$ around a certain value of $x$ means that the neural representation is stretched at that location. For $\mathbf{x}$ in dimension $K>1$, it is the $K\times K$ Fisher information matrix $F_{\text{code}}(\mathbf{x})$  which defines the local, non isotropic, geometry, $D(\mathbf{x},\mathbf{x}+\delta \mathbf{x}) \propto \;\delta\mathbf{x}^T\,F(\mathbf{x})\,\delta\mathbf{x}$. \\

The Fisher information $F_{\text{code}}(x)$ also characterizes the local neural geometry from the viewpoint of parameter estimation. The inverse of the Fisher information $F_{\text{code}}(x)$ is an optimal lower bound on the variance $\sigma_x^2 $ of any unbiased estimator $\widehat{x}(\mathbf{r})$ of $x$ from the noisy neural activity $\mathbf{r}$ \citep[Cramér-Rao bound, see e.g.][]{Blahut_1987}:
\begin{equation}
\sigma_x^2 \equiv \int \,  \big(\widehat{x}(\mathbf{r}) - x\big)^2 \;P(\mathbf{r}|x)\,d\mathbf{r}\;\geq \; \frac{1}{F_{\text{code}}(x)}
\label{eq:cramer_rao}
\end{equation}
Given that after category learning the Fisher information $F_{\text{code}}(x)$ is greater between categories, the Cramér-Rao bound implies that the variance in the estimation of an input $x$ is smaller between categories than within. This between category vs. within category difference means that near a category boundary, we expect that the variance should be lower in the direction of the boundary vs. parallel to it, where the change is within category, i.e., the variance will be anisotropic (see also Section~\ref{sec:coding_efficiency}).

\subsection{Neuronal noise as virtual new inputs} 
\label{sec:augmentation}
In biologically motivated models, neural noise is ubiquitous. Bayesian and information theoretic tools are well adapted to the study of such stochastic neural systems. In the simplest setting, neural spiking activity is described by a Poisson process. Hence, at each instant of time, in a feedforward architecture the input pattern from a layer to the next one has `missing data'. If we consider rates instead of spikes, one has an activity with a multiplicative noise, such as Poisson noise, very much in the same way as during a run of the dropout heuristic \citep{srivastava2014dropout}. Dropout is a widely used learning regularization technique consisting in perturbing the neural activity in any given layer with multiplicative Bernoulli noise, but other types of noise work just as well \citep[such as multiplicative Gaussian noise, also discussed in][]{srivastava2014dropout}. \citet{bouthillier2015dropout} propose an interesting interpretation of dropout: it serves as a kind of data augmentation \citep[see also][]{zhao2019equivalence}. In this work, the authors transform dropout noise into new samples: for a given input $x$, and a given perturbed neural activity $\mathbf{r}$, they compute the estimate $\widehat{x}$ that, in the absence of noise, would have produced an activity as close as possible to $\mathbf{r}$. In other words, presenting many times the same input to the noisy network is somewhat equivalent to presenting new inputs to a noiseless version of this network. This leads to an augmented training dataset, containing many  more samples than the original dataset. They show that training a deterministic network on this augmented dataset leads to results on par with the dropout results. We have seen in the previous section \ref{sec:geometry} that, given the local geometry characterized by the Fisher information $F_{\text{code}}$, the variance in the estimation of an input is smaller between categories than within.  Put in terms of data augmentation, regions that are within-category allow for more variability, whereas within cross-category regions, the acceptable variability is much more constrained in order to avoid generating a new input with a incorrect label  (\textit{i.e.} a virtual input crossing the decision boundary; see also \ref{sec:appendix_model}, section~\ref{sec:learning-cp} and Fig.~\ref{fig:posterior_estimation}). 

\subsection{A framework for the study of categorization in artificial neural networks} 
\label{sec:framework}

In the following, we study the building of categorical information in artificial neural networks making numerical experiments with two main techniques guided by the above formal analysis.\\

First, we consider a protocol inspired by the categorical perception experiments mentioned above. We assess the discrimination ability of each layer of an artificial neural network along a continuum of stimuli by considering the distance in neural space between contiguous elements. More distant stimuli are easier to discriminate than closer ones. We make use of a neural distance (the cosine distance between activities, see Materials and Methods), chosen as a proxy for the Fisher information $F_{\text{code}}$. This distance similarly quantifies how much the neural activity changes in average with respect to small variations in the input $x$.  However, contrary to Fisher information, it is straightforward to compute. Note that this quantity reflects sensitivity at the population level (a given layer in this study), and not at a single neuron level. In parallel to looking at the neural distance between neighboring stimuli along the continuum, in the simplest toy examples, we also compute for each stimulus in the continuum the variance of the virtual inputs as defined in the previous section.\\

Second, we consider the measure of a \textit{categoricality} index. Authors have proposed different measures of categoricality, either at the single neuron level \citep{kreiman2000category, freedman2001categorical}, or at the population level \citep{kriegeskorte2008matching, kreiman2000category}. The higher this index, the greater the intra class similarity and inter class dissimilarity. Here, we want to quantify categoricality at the population level. To do so, our choice of categoricality index consists in comparing the distributions of the distances in neural space, as given by the activations in each hidden layer, between items drawn from a same category vs. items drawn from two different categories (see Materials and Methods, paragraph ``Categoricality index'', for details). In what follows, this measure is computed from a held-out set different from the training set. 

\section{Results}
\label{sec:results}
We present the results of numerical experiments on classification tasks of increasing difficulty, working with a variety of datasets, from a simple one-dimensional example with two categories to a case that involves natural images with one thousand categories. We consider neural network architectures of complexity congruent with the ones of the tasks, allowing us to explore various conditions: multi-layer perceptrons and convolutional networks, a wide range of depths (from one hidden layers to ten or more hidden layers), and learning with different types of multiplicative noise (Bernoulli noise as in dropout, or Gaussian noise as in Gaussian dropout).

\subsection{Experiments with toy examples} 
\label{sec:toy}
\subsubsection{One dimensional example}
\label{sec:1d}
\begin{figure}[ht]
	\centering
	\includegraphics[width=\linewidth]{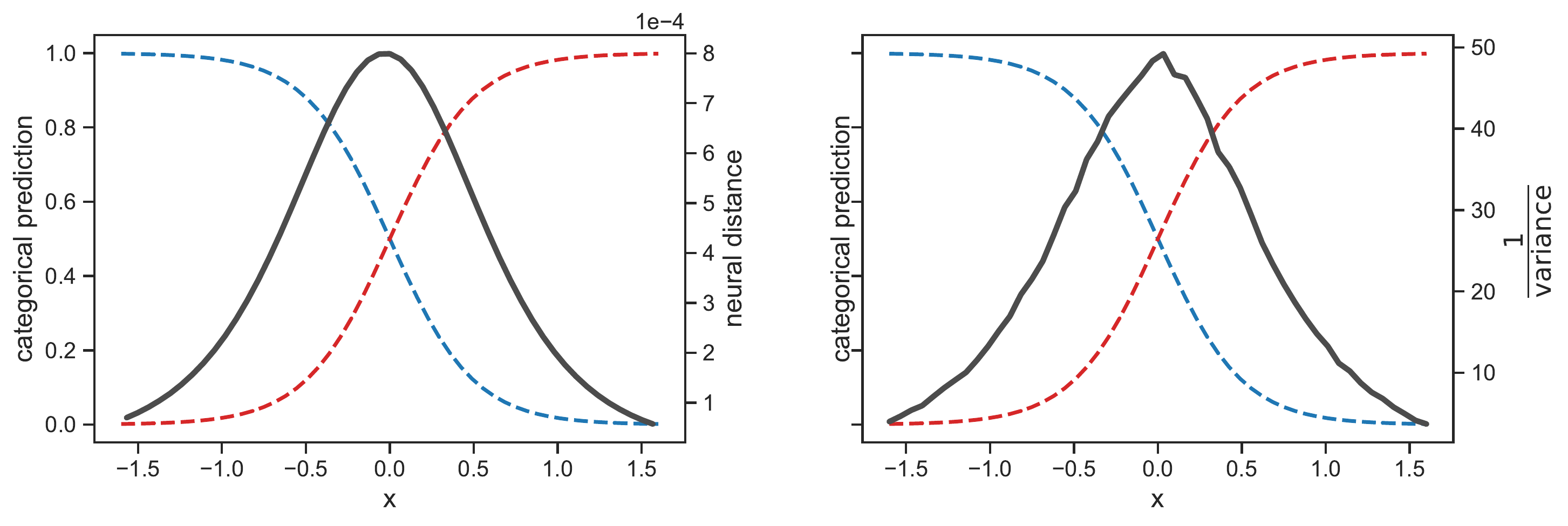}
	\caption{\textbf{One dimensional example with two Gaussian categories}, respectively centered in $x_{\mu_1} = -0.5$ and $x_{\mu_2} = +0.5$, with variance equal to $0.25$. For both panels, the dotted colored lines indicate the true posterior probabilities $P(\mu|x)$.
	(Left) The dark solid line corresponds to the distance in the neural space between contiguous stimuli.
	(Right) The dark solid line corresponds to one over the variance of the distribution of the estimated input $\widehat{x}(\mathbf{r})$ for $n=10000$ probabilistic realizations of the neural activity $\mathbf{r}$ given an input $x$. } 
	\label{fig:gaussian1d}
\end{figure}

We consider a one dimensional input space with two overlapping Gaussian categories. The resulting ensemble of stimuli corresponds to a continuum interpolating from a stimulus well inside a category to a stimulus well inside the other category, with a category boundary in between. The neural network is a multi-layer perceptron with one hidden layer of $128$ cells, with sigmoid activation, subject to Gaussian dropout with rate $0.5$ -- i.e. multiplicative Gaussian noise with standard deviation $1.0$. First, in Figure~\ref{fig:gaussian1d}, left panel, we present the neural distance, after learning, between contiguous inputs that are evenly distributed in stimulus space. As expected (see Section \ref{sec:CP_mod}), the network exhibits categorical perception, with greater distance in neural space between categories than within. Second, we generate virtual inputs as explained Section~\ref{sec:augmentation}:  for each considered stimulus $x$, we compute inputs which would have produced, in the absence of noise, an activity as close as possible to the neural activity evoked in the hidden layer by $x$ (see the Materials and Methods section for details on this computation). We present in Figure~\ref{fig:gaussian1d}, right panel, the inverse of the variance of the generated data-points at each point along the $x$ continuum after learning. In compliance with our analysis presented in Sections~\ref{sec:geometry} and~\ref{sec:augmentation}, this quantity is greater at the boundary between categories, paralleling the behavior of the distance shown in the left panel.

\subsubsection{Two dimensional example}
\label{sec:2d}
\begin{figure}[h]
	\centering
	\includegraphics[width=\linewidth]{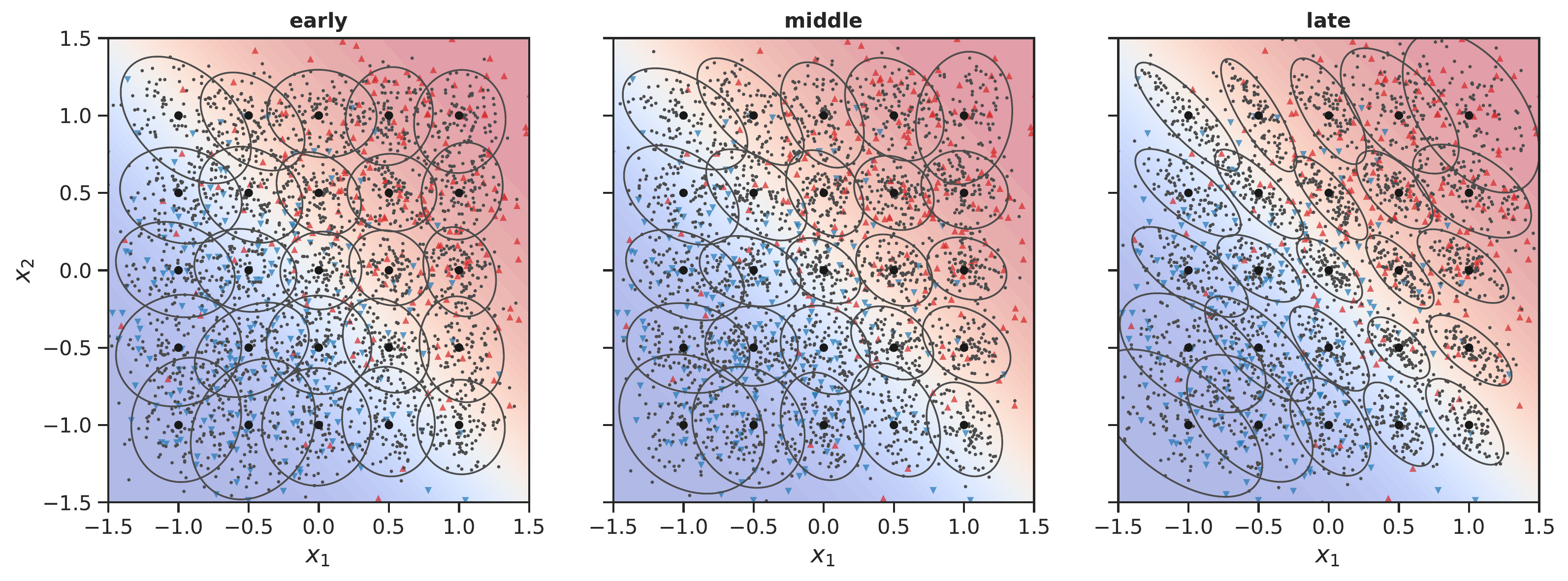}
	\caption[]{\textbf{Two dimensional example with two Gaussian categories}, respectively centered in $\mathbf{x}_{\mu_1} = \begin{psmallmatrix}-0.5\\-0.5\end{psmallmatrix}$ and $\mathbf{x}_{\mu_2} = \begin{psmallmatrix}0.5\\0.5\end{psmallmatrix}$ , with covariance matrix 
	$\mathbf{\Sigma} = \begin{psmallmatrix}0.25 & 0\\ 0 & 0.25\end{psmallmatrix}$.
	This example shows how the representation changes over the course of learning. We present snapshots taken at three points during learning: early (after 1 epoch), intermediate (after 4 epochs) and late (after 20 epochs).
	Individual training examples drawn from these multivariate Gaussian distributions are indicated as downward blue triangles (category 1) and upward red triangles (category 2). The background color indicates the true posterior probabilities $P(\mu|\mathbf{x})$, from blue (category 1) to red (category 2) through white (region between categories). The largest dark dots correspond to a $5\times5$ grid of stimuli tiling the input space between -1.0 and 1.0 in both dimensions. For each one of these inputs, we computed the estimates $\widehat{\mathbf{x}}(\mathbf{r})$ for $n=100$ probabilistic realizations of the neural activity $\mathbf{r}$. We represent these estimates as smaller gray dots, enclosed for each input by an ellipse that represents the $2\sigma$ confidence ellipse.} 
	\label{fig:gaussian2d}
\end{figure}

Similarly, we consider a two dimensional example with two Gaussian categories. The neural network is a multi-layer perceptron with one hidden layer of $128$ cells, with ReLU activation, subject to dropout with rate $0.2$ (proportion of the input units to drop, ie multiplicative noise drawn from Bernoulli distribution with $p=0.2$). As for the 1d case, we generate virtual inputs, here for a set of stimuli tiling the 2d input space. In Figure~\ref{fig:gaussian2d}, we see how the individual estimates (the generated virtual inputs) $\widehat{\mathbf{x}}(\mathbf{r})$ are distributed around each input $\mathbf{x}$. As expected, we observe that over the course of learning, the variance of these estimates for a given $\mathbf{x}$ gets lower near the boundary between the two categories, and that it is not isotropic: it is larger in the direction that is safe from crossing the boundary, and much smaller in the direction that is orthogonal to the decision boundary.

\subsection{Experiments with handwritten digits}
\label{sec:mnist}
In this section we move beyond the two simple previous examples to look at the MNIST dataset \citep{lecun1998gradient}. It is a database of handwritten digits that is commonly used in machine learning, with a training set of 60,000 images and a test set of 10,000 images. Here the goal is to look at categorical perception effects in neural networks that learn to classify these digits.

\subsubsection{Creation of an image continuum} 
\label{sec:mnist_continuum}
We want to build sequences of images that smoothly interpolate between two items of different categories. Directly mixing two stimuli in input space cannot work, as it would just superimpose the two original images. In cognitive experiments, one builds such artificial stimuli keeping each new stimulus as a plausible stimulus for our perception, as in the cases mentioned Section \ref{sec:CP}. One may think of various methods to generate sequences of images. In the present work, we want to obtain sequences that somehow remain in the space generated by the database itself. Taking inspiration from the work of \citet{bengio2013better} to create an image continuum, we used an autoencoder trained to reproduce single digits from the MNIST training set (see Materials and Methods for algorithmic details). The autoencoder consists in an encoder and a decoder. The first part learns to build a compressed representation of the input. The second part learns to reconstruct the original input from this compressed representation. This representation projects the data onto a lower dimensional space that meaningfully represents the structure of the data. By interpolating between two stimuli in this space, then reconstructing the resulting image thanks to the decoder, we obtain a continuum that nicely morphs between two stimuli, along the nonlinear manifold represented by the data.\\

\begin{figure}
	\centering
	\includegraphics[width=0.99\linewidth]{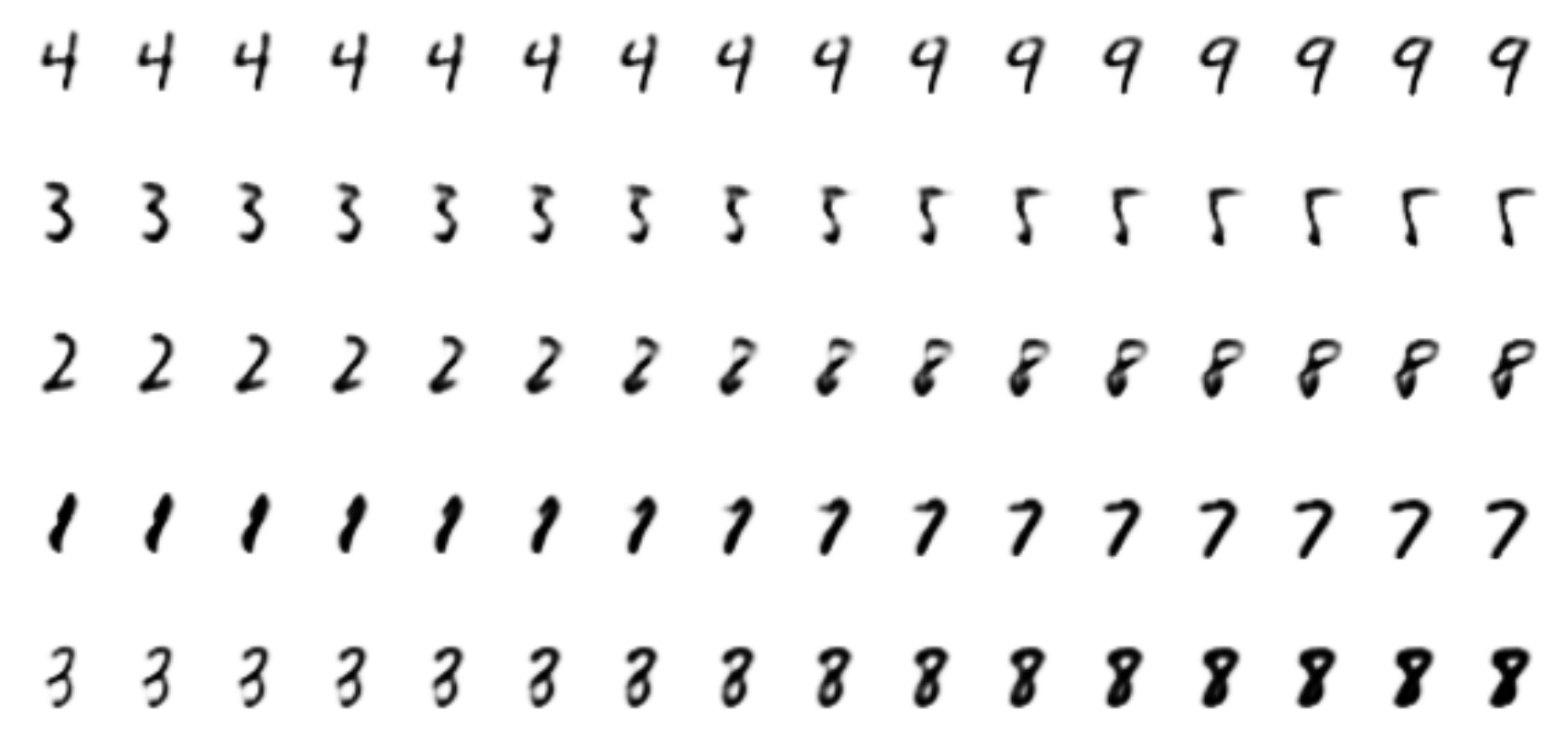}\\
	\caption{\textbf{Examples of image continua}, smoothly interpolating between pairs of digits taken from the MNIST test set.}
	\label{fig:mnist_continua}
\end{figure}

We provide in Figure~\ref{fig:mnist_continua} several examples of such generated continua, each one interpolating between two digits from the MNIST test set. 

\subsubsection{Peak in discrimination at the boundary between categories}
\label{sec:mnist_cp}
\begin{figure}
	\centering
	\includegraphics[width=0.48\linewidth]{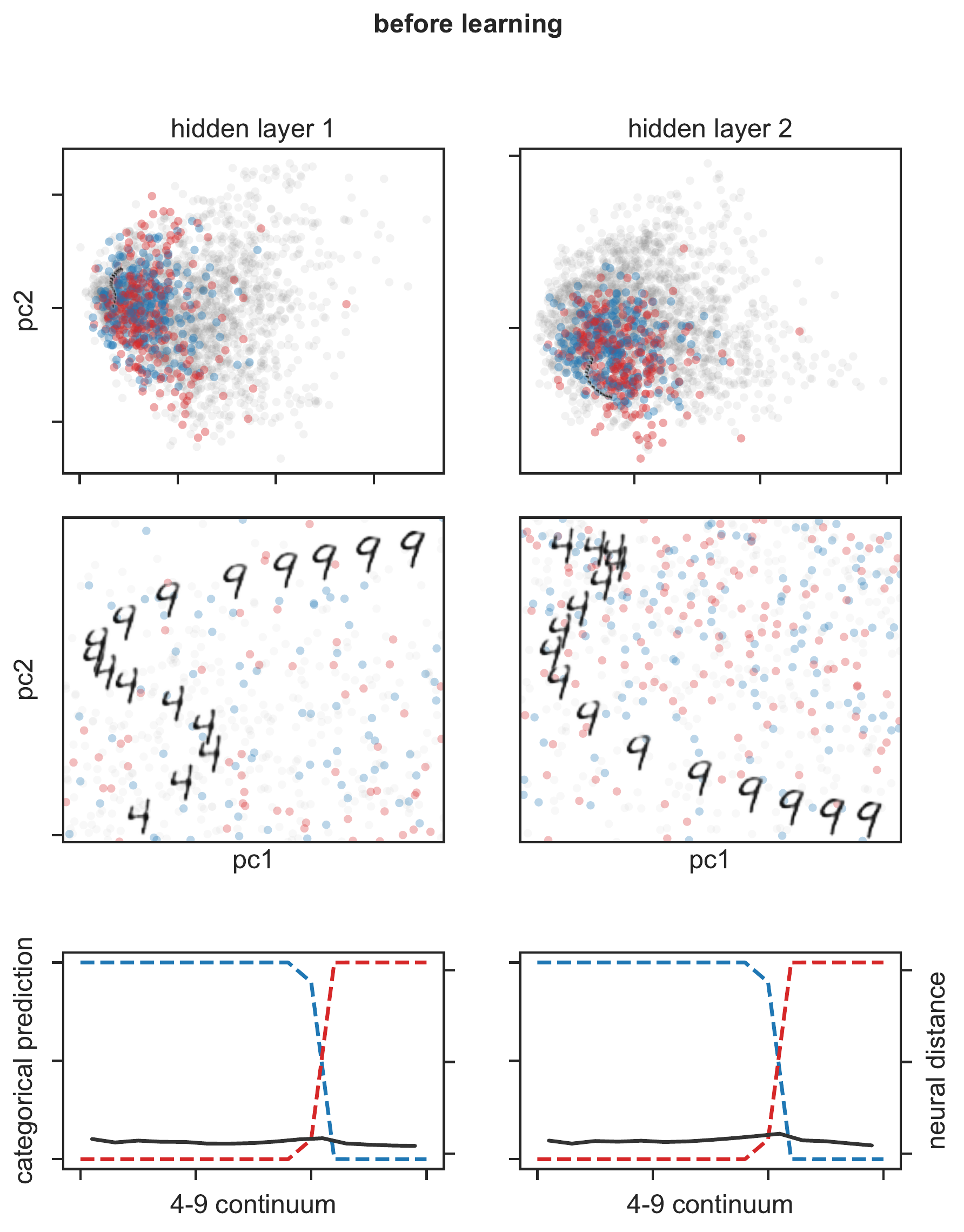}
	\hfill
	\includegraphics[width=0.48\linewidth]{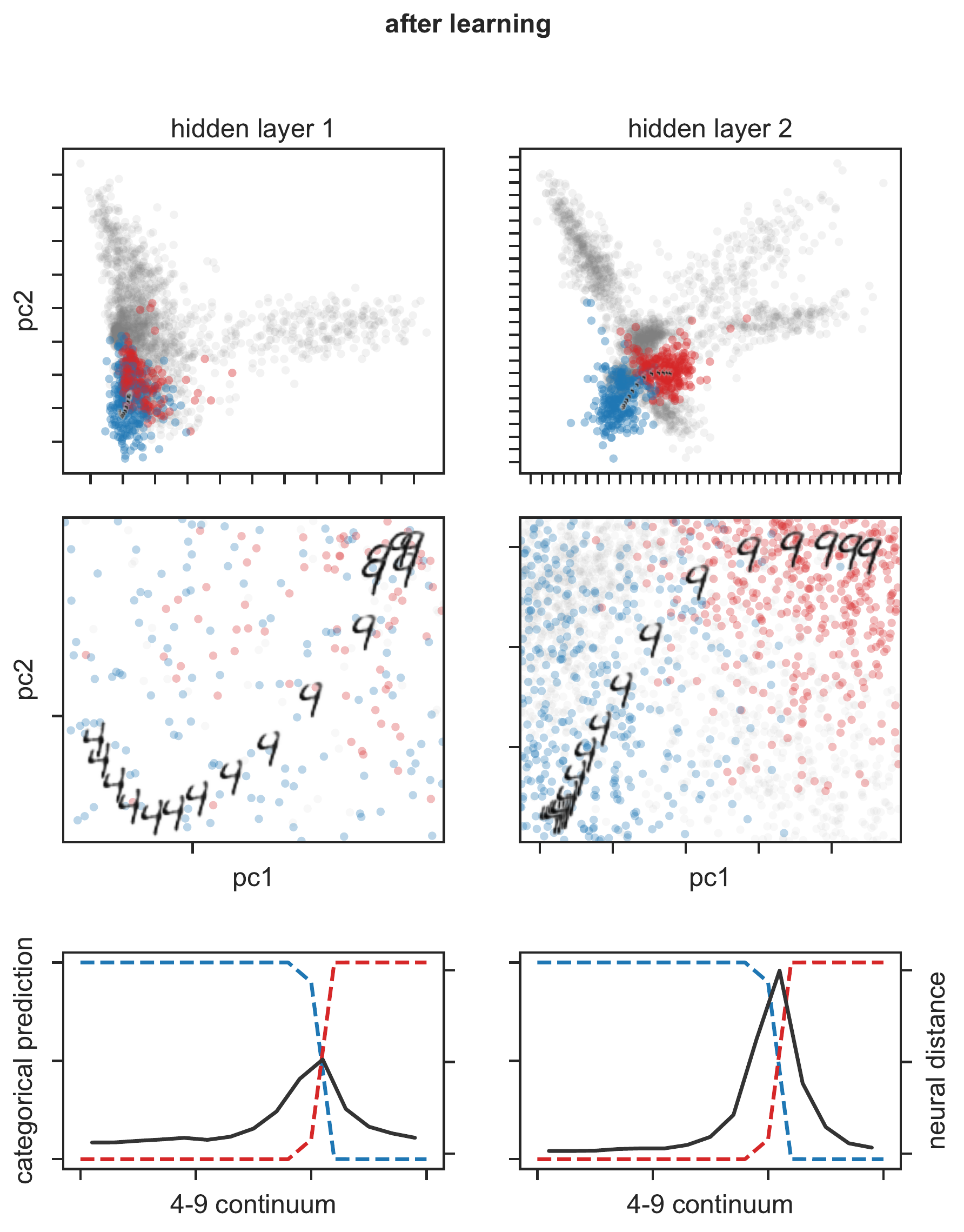}
	\caption{\textbf{Changes in the neural representation following learning of categories}: example on a `4' to `9' continuum, using the MNIST dataset. The neural network is a multi-layer perceptron with two hidden layers of 256 cells. (Left) Representation before learning. (Right) Representation after learning. (Top row) Two-dimensional PCA projections based on the activations of the hidden layers on the test set. Items from category `4' are colored in blue, while items from category `9' are colored in red. The rest of the test set is represented in gray. For a better visualization, only one every four data points is shown. One specific `4' to `9' continuum, connecting two items from the test set is represented in black. (Middle row) Same, zoomed in on the `4' to `9' continuum. (Bottom row) The dotted colored lines indicate the posterior probabilities, as found by the network, of category `4' (blue) or `9' (red) along the continuum. The dark solid line indicates the neural distance between adjacent items along the continuum. The scale along the y-axis is shared across conditions.}
	\label{fig:mnist_cp}
\end{figure}

In this section we consider a multi-layer perceptron with two hidden layers of 256 cells, and look at the changes in representation before and after learning on the whole MNIST training set. We investigate the behavior of the network with respect to the 4--9 continuum presented on top of Fig.~\ref{fig:mnist_continua}. The 4/9 classes are among the most confused ones, hence of particular interest for the present study which focuses on cases where the stimuli can be ambiguous. We summarize the results in Figure~\ref{fig:mnist_cp}, for each of the two hidden layers, before and after learning. Before training (left panel), the representation of all $4$ and $9$ digits are not well separated. If one looks at the neural distance between items along the 4--9 continuum (bottom row), we see that it is rather flat. Conversely, after training (right panel), the two categories are now well separated in neural space. The neural distance between items along the specific 4--9 continuum presents a clear peak at the decision boundary, thus exhibiting categorical perception. Again, this is predicted by the analysis presented in Section~\ref{sec:CP_mod}: as the category uncertainty increases, the neural distance increases. We can also already notice that the deeper hidden layer exhibits a more categorical representation.

\subsubsection{Gradient categorical perception as a function of depth}
\label{sec:mnist_depth}
\begin{figure}
	\centering
	\includegraphics[width=0.5\linewidth,valign=T]{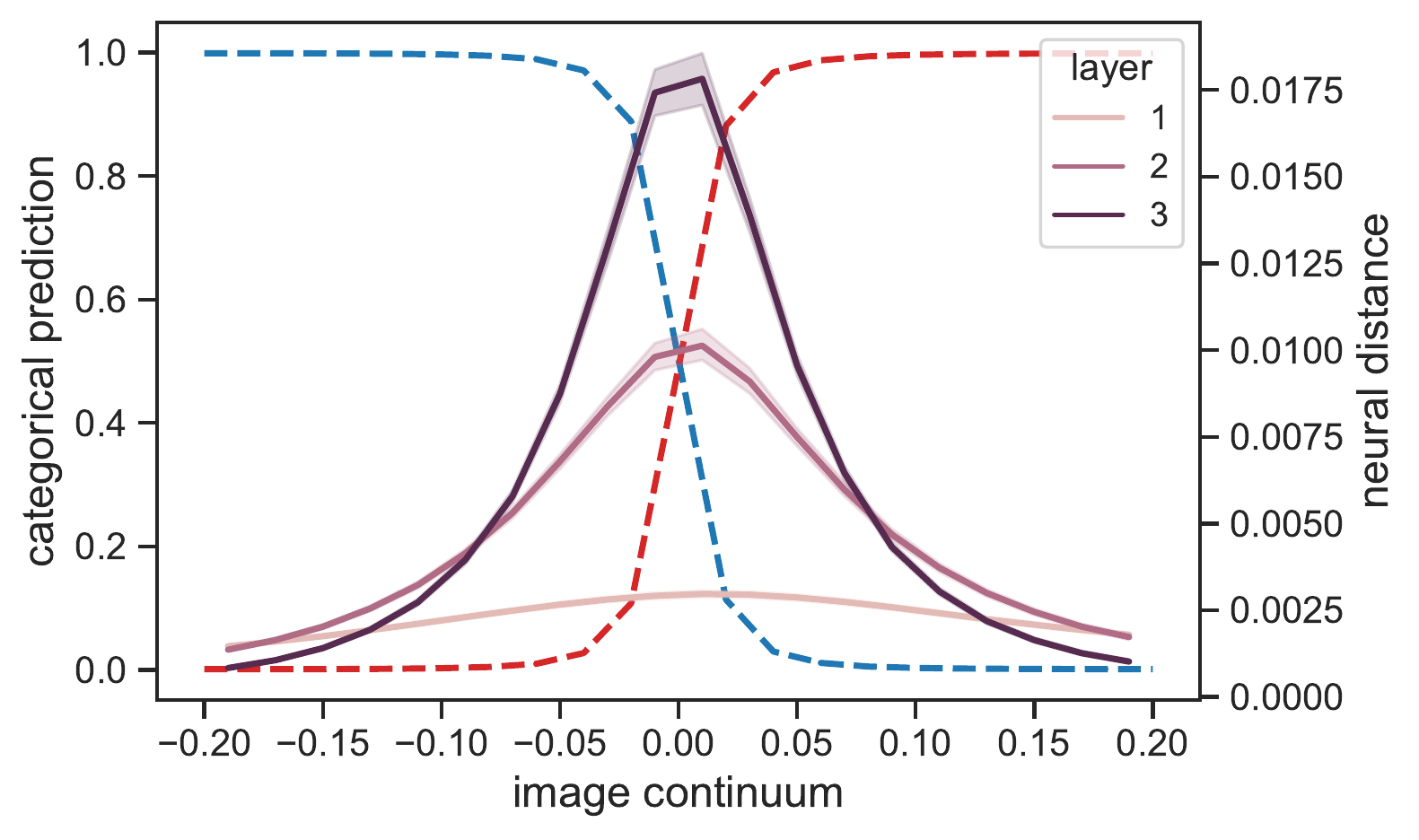}
	\caption{\textbf{Gradual categorical perception across layers}: the deeper the layer, the more pronounced the categorical perception effect. The neural network is a multi-layer perceptron with three hidden layers of 256 cells, trained on the whole MNIST dataset. The dotted colored lines indicate the mean posterior probabilities from the network. 
	For each hidden layer, the solid line corresponds to the mean neural distance between adjacent items, averaged over several continua, and aligned with the boundary between the two classes (error bars indicate 95\% confidence intervals, estimated by bootstrap).}
	\label{fig:mnist_cp_depth}
\end{figure}

After having studied the properties along a single continuum, we now make a statistical analysis by looking at the pattern of discriminability along an ensemble of many similar continua that interpolate between pairs of stimuli from different categories (see Materials and Methods). We present in Fig.~\ref{fig:mnist_continua} a few examples of such continua. In order to look at the effect of depth, we consider here a multi-layer perceptron with three hidden layers, trained on the whole MNIST training set. For the particular examples shown in Fig.~\ref{fig:mnist_continua}, we plot in Fig.~\ref{fig:mnist_continua_examples_input_distance_pred} the labeling response provided by this network after learning, together with the distance in input space between adjacent items along each one of these continua. For each continuum in our large ensemble, we computed the neural distance between neighboring stimuli along the continuum, and this for each hidden layer. In order to compute averages over the set of continua, we aligned all these curves by centering them around the point where the two posterior probabilities cross. We present the results in Figure~\ref{fig:mnist_cp_depth}. We first observe that all layers exhibit categorical perception: space is dilated at the boundary between categories and warped within a category. Moreover, we see that the deeper the layer the more pronounced the effect.

\subsubsection{Categoricality as a function of depth and layer type}
\label{sec:mnist_categoricality}
We now turn to a second method for characterizing how much the neural code is specific to the categorization task, making use of the categoricality index mentioned Section \ref{sec:framework}. We recall that this index quantifies the degree of relative intra-class compression vs. inter-class expansion of the neural representation provided by a given layer (see Materials and Methods). A reasonable expectation is that after learning, categoricality increases with layer depth, since the input is not categorical, and the decision layer is constrained to be categorical. The issue now is to characterize how categoricality changes as a function of depth.\\

\begin{figure}
	\centering
	\textbf{a.}
	\includegraphics[width=0.44\linewidth,valign=T]{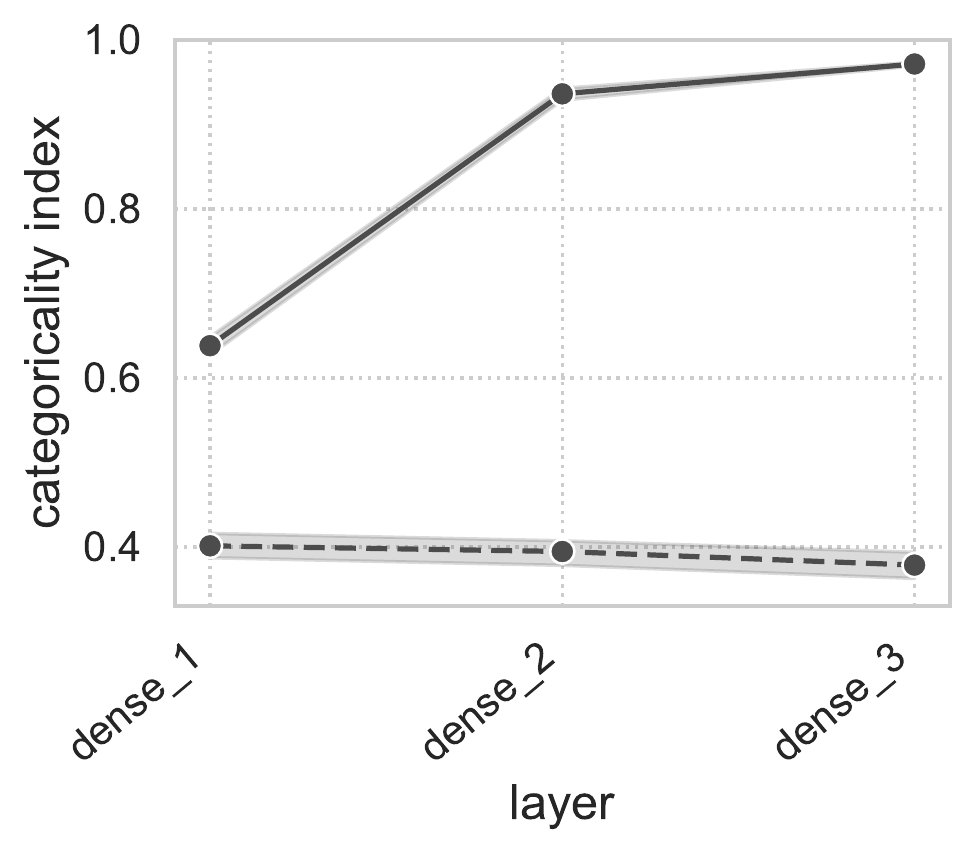}
	\hfill
	\textbf{b.}
	\includegraphics[width=0.44\linewidth,valign=T]{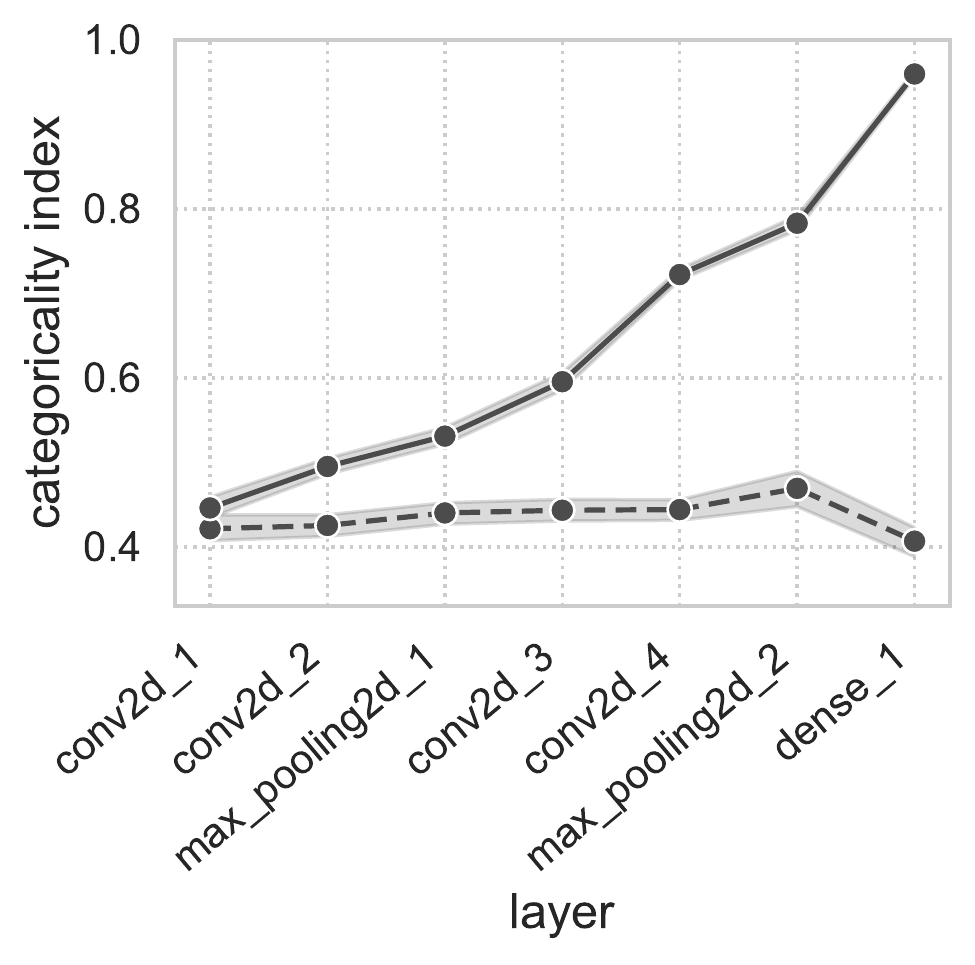}
	\caption{\textbf{Categoricality as a function of layer depth, using the MNIST dataset}. Categoricality is a measure that quantifies the distance between items drawn from the same category vs. different categories, thanks to the Kolmogorov-Smirnov statistic between the intra- and inter-category distributions of neural distances. Solid lines correspond to trained networks, broken lines to untrained ones (error bars indicate 95\% confidence intervals, estimated by bootstrap). (a) The neural network is a multi-layer perceptron with three hidden layers.
	(b)  The neural network is a convolutional neural network whose layer structure is described in the x-axis of the figure.}
	\label{fig:mnist_categoricality}
\end{figure}

In Figure~\ref{fig:mnist_categoricality} we compare the categoricality index before (broken line) and after (solid line) learning on the MNIST training set for two types of neural network: on the left, (a), a multi-layer perceptron with three hidden layers, and on the right (b), a convolutional neural network with seven hidden layers (see Materials and Methods for full details). Let us first consider the multi-layer perceptron. The results in Fig.~\ref{fig:mnist_categoricality}a echo the one presented in Fig.~\ref{fig:mnist_cp_depth}: all layers present categorical effects, and, in line with recent findings \citep{alain2016understanding, mehrer2020individual},  categoricality increases with depth. Let us now turn to the convolutional neural network. The first convolutional layer does not have a larger categoricality index than an untrained network, meaning that the features it has learned are quite general and not yet specialized for the classification task at hand. The categoricality then increases with depth, the last hidden layer, a dense layer, presenting the largest value. Finally, comparing the two figures, we can see that the dense layers in the fully-connected multi-layer perceptron on the left have a larger categoricality index than the convolutional layers, even for the first layer.

\subsection{Experiments with natural images}
\label{sec:naturalimage}
We now go one step further in task and network complexity by considering natural image databases and deeper networks with ten or more hidden layers.

\subsubsection{Categorical perception of a cat/dog continuum}
\label{sec:cat_dog}
\begin{figure}
	\textbf{a.}
	\includegraphics[width=.95\linewidth,valign=T]{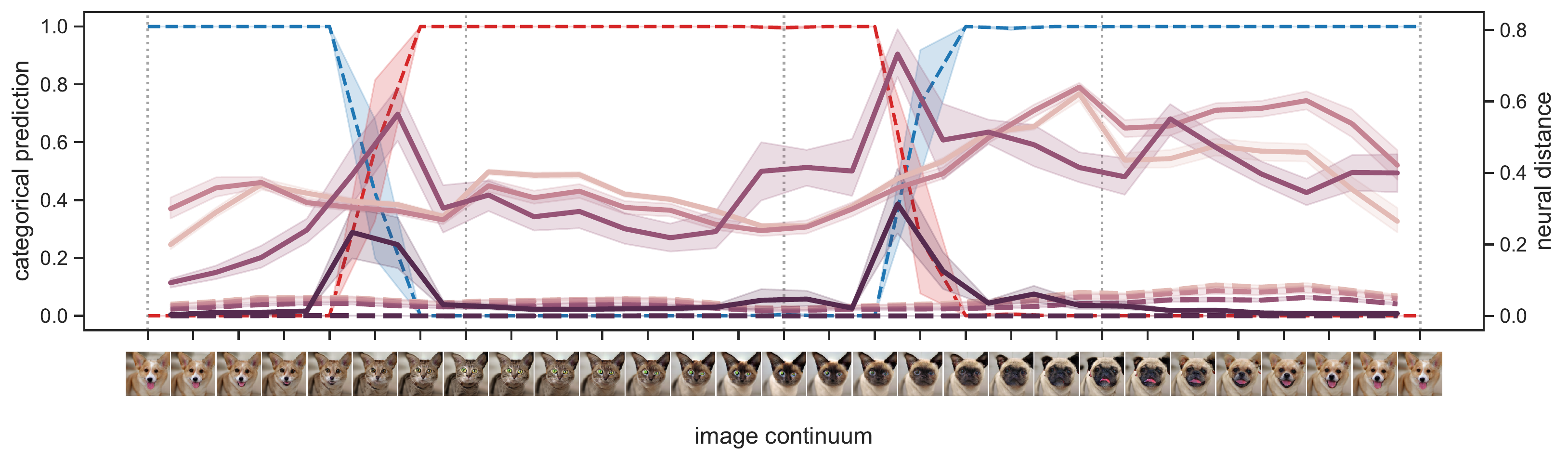}
	\vspace{0.1cm}\\
	\textbf{b.}
	\includegraphics[width=.95\linewidth,valign=T]{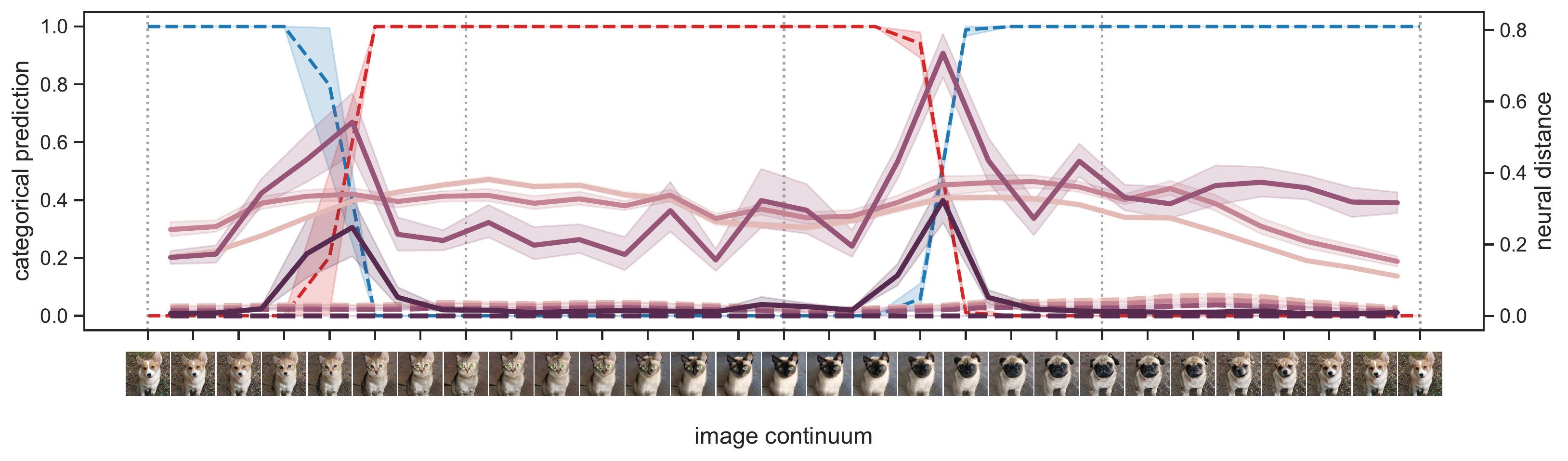}
	\caption{\textbf{Categorical perception of a cat/dog circular continuum}. Experiment with continua interpolating between cats and dogs, with two different viewpoints: (a) close up on the face, and (b) full body. 
	The interpolations involve two types of dogs and two types of cats. Each continuum corresponds to a circular interpolation with four sub-continua: from the first dog to a cat, then to the other cat, then to the second dog, and finally back to the first dog. The neural network is a convolutional neural network with three blocks of two convolutional layers and a max pooling layer, finally followed by a global average pooling layer. The blue and red dashed lines indicate the posterior probabilities from the network (blue is dog, red is cat). The colored solid lines correspond to the neural distance between adjacent items along the continuum, the darker the line the deeper the layer. Only the last convolution layer of each block and the global average pooling layer are shown. The colored dotted lines are the counterparts for the same networks but before learning. Error bars indicate 95\% confidence intervals, estimated by bootstrap. The thin dotted vertical lines indicate the start and end points of each sub-continua.} 
	\label{fig:cat_dog}
\end{figure}

In this section, we consider a deep convolutional neural network trained to classify natural images of cats and dogs. We investigate its behavior with respect to a continuum that interpolates between different cat/dog categories. Let us first introduce the neural network, the database used for training and finally the continua that are considered (see Materials and Methods for details). The neural network is a convolutional network with three blocks of two convolutional layers and a max pooling layer, followed by a global average pooling layer. We used Gaussian dropout during learning. We trained each network instance on the Kaggle Dogs vs. Cats database, which contains 25,000 images of dogs and cats, with a final classification performance on the test set of about 95\%. In order to assess the changes in representation induced by learning for the different layers in the neural network, we  considered different continua that either interpolate between items from the two categories, in which case we expect categorical perception to emerge, and between items from the same categories, as a control. The continua that we consider cycle from one dog to the same dog, going from one dog to a cat to another cat to another dog and finally back to the first dog. Each sub-continuum is made of 8 images, thus totaling 28 images for a given full continuum. We considered two such continua, using the same cat/dog categories but considering different viewpoints (close up vs full body -- see the x-axes of Fig.~\ref{fig:cat_dog}a and b for the morphed images that were used as input). The different cat/dog, cat/cat or dog/dog morphed continua were generated thanks to the code and pretrained model provided by \citet{miyato2018spectral} that uses a method based on Generative Adversarial Networks \citep{goodfellow2014generative}. Note that these continua on which the neural networks are tested have been generated by a network trained on a completely different database.\\

We present the results in Figure~\ref{fig:cat_dog}. First, one can see that the network well categorizes the different cat and dog images (see the dotted blue and red lines) into the correct classes. The last hidden layer, right before the final decision, exhibits a strong categorical perception effect (see the darkest line): only items that straddle the categories can be discriminated. In contrast, the first convolutional layers do not exhibit categorical perception: there is no clear peak of discrimination between categories. Instead, differences between contiguous images appear to mainly reflect differences in input space (as a comparison, see \ref{sec:appendix_catdog_pixel},  Fig.~\ref{fig:cat_dog_input} for a picture of the distances in input space). For instance, the peak difference in input space and for these first convolutional layers is driven by the tongue sticking out of the mouth, which does not affect the more categorical upper layers. Finally, the last convolutional layers exhibit in-between behavior, with an ability to discriminate between within-category stimuli, but with a clear peak at the cat/dog boundary, thus displaying categorical perception. 

\subsubsection{Categoricality in deep networks}
\label{sec:imagenet}
In this section, we work with the ImageNet dataset \citep{deng2009imagenet}, and more precisely with the subset of images used in the ILSVRC-2010 challenge \citep{ILSVRC15}. The network that we consider is the VGG16 model described in \citet{simonyan2014very}, which has won the ImageNet Challenge 2014. This model is characterized by 16 weight layers, an architecture considered very deep (at the time this VGG16 model was published). Here, we compare the categoricality index on randomly initialized networks with the exact same architecture, and on a network that has been pretrained on the ImageNet database (as provided by the keras package\footnote{\url{https://keras.io/api/applications/vgg/}}) (see Materials and Methods for details).\\

\begin{figure}
	\centering
	\includegraphics[width=.8\linewidth]{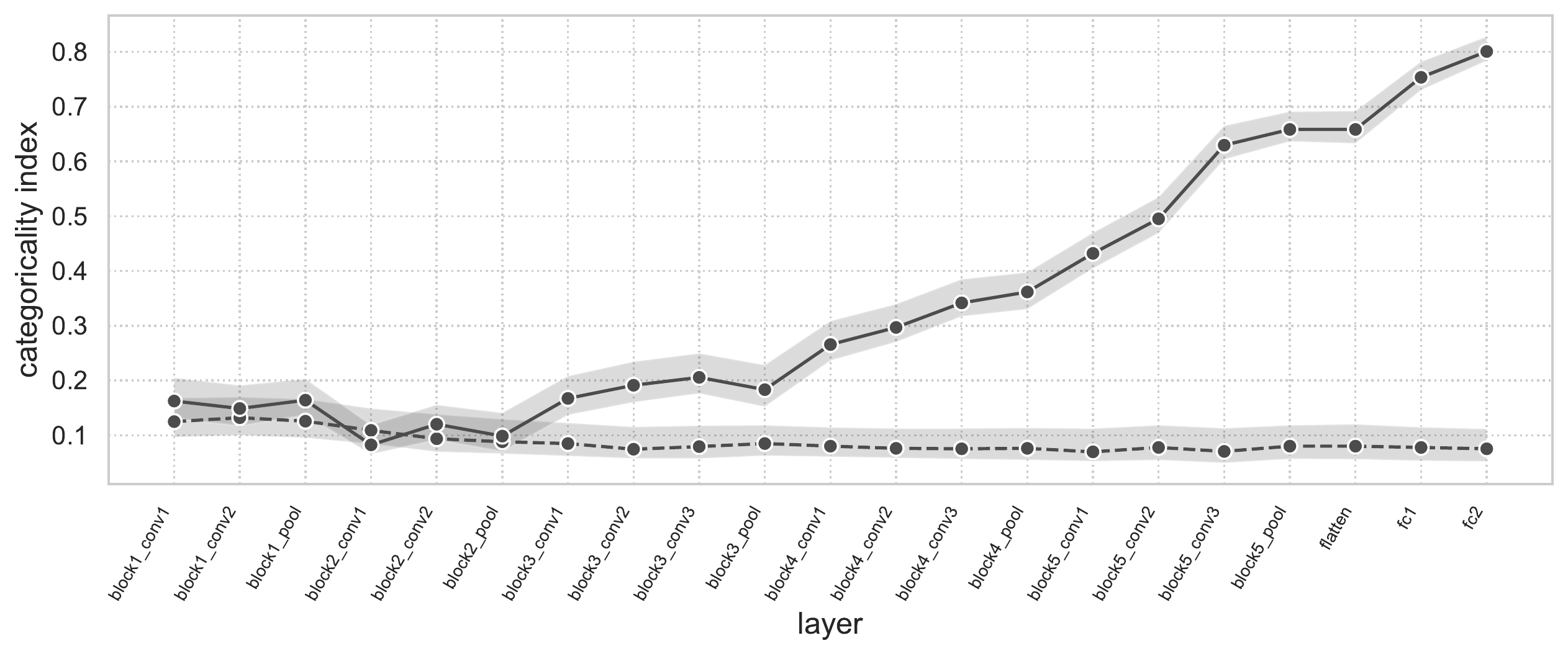}
	\caption{\textbf{Categoricality as a function of layer depth, using the ImageNet dataset}. The neural network is the VGG16 model \citep{simonyan2014very}. The dash line corresponds to a network with random initialization of the weights, whereas the solid line corresponds to a network pretrained on ImageNet (error bars indicate 95\% confidence intervals, estimated by bootstrap).} 
	\label{fig:imagenet_categoricality}
\end{figure}

We show in Figure~\ref{fig:imagenet_categoricality} the results of this comparison. As expected, one can see that for an untrained network, the categoricality index is flat across layers: the neuronal layers do not show any preferential knowledge of the categories. For a trained network, as seen in the MNIST section above, the categoricality increases as a function of depth. We can observe that, for the first convolutional layers, this index is essentially not much different from the case of an untrained network. Intermediate convolutional layers do exhibit some effects of category learning, while for the last convolutional layers the categoricality is much more marked. Finally, the last two dense layers exhibit the greatest categoricality.

\section{Discussion}
\label{sec:discussion}

\subsection{A new view of dropout} 
\label{sec:dropout}
In the present discussion we show how our formal and numerical analyses of categoricality in multi-layer networks help elucidating various empirical observations and heuristic practices in the use of the dropout technique. As we have seen, the effect of learning categories leads to neural representations with a distorted geometry: a finer resolution is obtained near a class boundary, the Fisher information of the neural code becoming greater near the boundary between categories than within categories. Experimentally, we showed that after learning, neural distance between neighboring stimuli is indeed greater in these boundary regions compared to regions well within a category. As discussed, this categorical effect builds up gradually in the course of learning, becoming stronger for deeper layers, and stronger for dense layers compared to convolutional ones. These geometrical properties have the consequence of controlling the impact of neuronal noise as a function of the probability of misclassification (see Section~\ref{sec:CP_mod} and \ref{sec:appendix_model}). Thus the level of acceptable noise depends on how much the neural geometry in the considered layer is adapted to the category geometry. It is thus different at different learning stages, and different depending on layer depth and layer type. Dropout precisely injects noise in each layer, opening the possibility of a noise level adapted locally with respect to the current (in time and depth) congruence of the neural representation with the category geometry. We speculate here that this very possibility of having the noise (dropout) level adapted to the current neural geometry induces a positive interaction between noise and learning.\\

\citet{bouthillier2015dropout} propose that dropout is equivalent to a data augmentation technique, in that injecting such noise would be equivalent to processing additional stimuli (inputs). As discussed Section~\ref{sec:augmentation}, this data augmentation is not uniform with respect to stimulus space: the generated stimuli, following the method proposed by \citet{bouthillier2015dropout}, do not distribute uniformly over the input space (see Fig.~\ref{fig:gaussian1d} and Fig.~\ref{fig:gaussian2d}). Their variability is adapted to the structure of the categories, being greater within a category than between categories, where the risk of generating a new stimulus with a wrong label is highest. Hence, if one considers dropout as a data augmentation technique~\citep{bouthillier2015dropout,zhao2019equivalence}, injecting noise in the hidden layers is likely to yield better results than techniques that only consider input noise, as this latter type of noise might be too weak within a category and too strong between categories.\\

From this viewpoint, we expect that it should be beneficial to inject more dropout when and where it is allowed to have more noise -- hence at later stages of learning, in deeper layers, in dense layers vs. convolutional layers. If all this is correct,  we then have a coherent picture which means that there is a positive interaction between categoricality and dropout: more categoricality allows for more noise, allowing to better benefit from the data augmentation, which in turn helps increasing categoricality, inducing a greater separation between categories. We now show that the widespread practices in the use of dropout, together with supplementary numerical simulations, support such conclusion. Although more work is needed to quantitatively investigate the effectiveness of dropout, it is reasonable to assume that the most common practices reflect intuitions and trial and error explorations leading to the selection of beneficial heuristics.

\paragraph{More dropout in deeper layers.} 
A widespread practice is to use greater dropout level for the hidden layers compared to the input layer (see \citealp{Goodfellow-et-al-2016}\footnote{``Typically, an input unit is included with probability 0.8, and a hidden unit is included with probability 0.5.''}). In their original dropout article, \citet{srivastava2014dropout} also use three levels of dropout rate, with noise increasing with the layer depth\footnote{see ``Dropout was applied to all the layers of the network with the probability of retaining a hidden unit being p = (0.9, 0.75, 0.75, 0.5, 0.5, 0.5) for the different layers of the network (going from input to convolutional layers to fully connected layers).''}. Yet, to our knowledge, there is no systematic study of the optimal dropout level as a function of the depth of the layer. In Appendix \ref{sec:appendix_dropout_depth}, we provide a first numerical investigation providing evidence that the optimal dropout rate increases with layer depth. We trained a  multi-layer perceptron on the CIFAR-10 image dataset \citep{krizhevsky2009learning}, with varying amount of dropout rate after the input layer and each of the two hidden layers. Fig.~\ref{fig:cifar10_dropout} presents the results. First, one can see that the deeper the layer the more robust it is to a great amount of noise, and, second, that the best dropout value for each layer increases with depth.

\paragraph{Categoricality increases without dropout, but dropout helps.}
\citet{mehrer2020individual} have found that an increase of the dropout probability yields greater categoricality (see their Fig. 8c). Our simulations in the Section \ref{sec:mnist_categoricality} made use of a dropout rate that increases with layer depth. One may then ask whether this use of dropout is not actually the main source in the difference of categoricality between layers.
We thus performed a control experiment on the MNIST dataset reproducing the results of Fig.~\ref{fig:mnist_categoricality} Section  \ref{sec:mnist_categoricality}, but without the use of dropout. The results are presented in Appendix \ref{sec:appendix_dropout_mnist},  Fig.~\ref{fig:mnist_categoricality_nodropout}. We find that categoricality does build up in the absence of dropout, and that after learning the categoricality index does increase as a function of depth. Yet, dropout helps increasing the separation between categories: the slope of the increase of the categoricality index when learning with (Bernoulli or Gaussian) dropout is slightly larger than without the use of noise.

\paragraph{Dense vs.~convolutional layers.} 
As we have shown in Sections \ref{sec:mnist_categoricality} and \ref{sec:imagenet}, dense layers exhibit more categoricality than their convolutional counterparts, and convolutional layers exhibit very little categoricality when close to the input layer, whereas the deepest ones do exhibit some categoricality. In agreement with our expectations, the dropout level is usually chosen to be stronger for dense layers, convolutional layers closest to the input layer receive a small amount of dropout or no dropout at all, and a gain of performance is obtained thanks to dropout in the deepest convolutional layers (\citealp{srivastava2014dropout}; see also \citealp{park2016analysis,spilsbury2019don}). 

\paragraph{Increasing the dropout rate during learning is preferable.}
Several works have considered adapting the dropout rate during the course of learning. Relying on the simulated annealing metaphor, \citet{rennie2014annealed} suggest decreasing the dropout rate over the course of training, starting from a high initial value. The intuition is that at first high noise makes it possible to explore the space of solutions, avoiding local minima, with a last phase of fine-tuning without noise. On the opposite side, taking inspiration from curriculum learning~\citep{bengio2009curriculum}, \citet{morerio2017curriculum} propose the exact contrary: to start learning with low noise, and then increase the difficulty by raising the amount of noise. As discussed, a network needs to have already partly learned the structure of the categories in order to allow for the use of a high level of dropout. We thus expect the curriculum dropout to have better performance than the original dropout (that is with a fixed value of the dropout rate over the course of training), and better than the annealed dropout: this is  precisely what is found in the experiments by \citet{morerio2017curriculum} on a variety of image classification datasets \citep[see also][]{spilsbury2019don}.

\paragraph{Nonlinear interaction with the size of the dataset.} 
Finally, from the interaction that we describe between learning and dropout, we also expect an interaction with the size of the dataset. Obviously, with a very large number of training samples, there is no need for a regularization technique such as dropout, as there is no worry about overfitting. Less intuitive is the expectation here that if the number of training samples is too small, since the structure of the categories cannot be sufficiently learned, the use of dropout will be detrimental to the performance. In between, we expect the introduction of noise to be beneficial to the generalization ability of the neural network. All in all, this is exactly what is found by \citet[][see Fig.~10]{srivastava2014dropout}.

\subsection{Relevance to the fields of psychology and neuroscience}
\label{sec:neuro}

\subsubsection{An historical debate on categorical perception}
Historically, categorical perception was first presented as an all-or-nothing phenomenon according to which subjects can discriminate between stimuli only through their phonetic labels -- or more precisely through their class identification probability \citep[following the view of the classical Haskins model; see][]{Liberman_etal_1957}. As a result, discrimination was thought to be almost zero within category. In our setting, it would be as if discrimination was only made possible through the use of the very last layer of the neural network, the one corresponding to the decision process with the softmax function. However, from the very beginning, this idealized form of categorical perception has never been met experimentally (see, e.g., \citealp{Liberman_etal_1957} or \citealp{liberman1961effect}; see \citealp{lane1965motor, Repp_1984} for reviews): observed discrimination is always above the one predicted from the identification function. Some authors rather talk about \textit{phoneme boundary effect} in order to distinguish it from the original categorical perception proposal \citep{wood1976discriminability, iverson2000perceptual}. As this phenomenon is not limited to speech, we keep here the term \textit{categorical perception}, simply characterized as a better ability to perceive differences between stimuli in the cross-category regions than within a category.\\

Our framework makes it possible to better understand this phenomenon. As we have seen, category learning does not only affect the last decisional layer, but also the upstream coding layers in order to better represent categories and robustly cope with noise. For each hidden layer, we observe a warping of the neural space within category and an expansion between categories. Thus, without the need to even actually compute labels, assuming the coding layers are reused during discrimination experiments, we expect to see categorical perception effects during discrimination tasks, but possibly with an above chance discrimination within category, as indeed found experimentally. This effect is not due to labeling, but a consequence of the optimization of the coding layers upstream of the categorization final process.\\

As we have seen, a deep network exhibits a gradient of categorical perception, from the first layers having almost no traces of categoricality to the last layers that show an important distortion of the space according to the structure of the categories. Where exactly is the basis for our conscious perception remains an important research question.

\subsubsection{Results in neuroscience and imagery}
Neurophysiological studies (\citealp{xin2019sensory}; \citealp[][see their Fig. 6 panel H]{okazawa2021thegeometry}) have recently shown that category learning leads to neural representations with a geometry in agreement with our prediction of expansion/contraction along a direction quantifying the degree of ambiguity of the stimulus category (see \citealp{LBG_JPN_2008}, and Section~\ref{sec:CP_mod}).\\

In the present work, we notably study categorical representations by looking at the neural distance between stimuli that are equally spaced in stimulus space. Previous experimental works have also compared stimuli in neural space by looking at the distance between activities. Using high‐density intracranial recordings in the human posterior superior temporal gyrus,  \citet{chang2010categorical} have found that this region responds categorically to a /ba/--/da/--/ga/ continuum (as used in the original \citealp{Liberman_etal_1957} study): stimuli from the same category yield indeed more similar neural patterns that stimuli that cross categories. Still in the domain of speech perception, using event-related brain potentials, \citet{bidelman2013tracing} followed a similar approach in comparing neural activities in response to a vowel continuum. They found that the brainstem encodes stimuli in a continuous way, with changes in activity mirroring changes in the sounds, contrary to late cortical activity that are shaped according to the categories. This is in line with the view  of gradient effects of categorical perception as function of depth in the processing stream: input layers show almost no effect of categories, whereas last layers are strongly affected by them.\\

The monkey study by \citet{freedman2003comparison} have shown a distinct behavior of the prefontal and inferior temporal cortices during a visual categorization task. The visual processing stream goes from sensory to the prefrontal cortex (PFC) through the inferior temporal cortex (ITC). In order to assess the categorical nature of the activity of each region, \citet{freedman2003comparison} introduced an index similar to what we use in our study, by comparing the responses to within- vs. between-categories pairs of stimuli, but at the level of an individual neuron. In agreement with the picture proposed here, the authors have found that both regions show significant effects of category learning, with larger differences in the neural responses for pairs of stimuli that are drawn from different categories, and that the PFC neurons show stronger category effects than the ITC neurons. Note though that the categoricality of the ITC is actually underevaluated due to the use of a measure at the single neuron level. A subsequent study by the same team has indeed shown that the ITC contains actually more categorical information when analyzed at the population level \citep[see][]{meyers2008dynamic}. Similarly, using both electrode recordings in monkeys and fMRI data in humans,  \citet{kriegeskorte2008matching} have shown that the ITC exhibits a categorical representation, contrary to the early visual cortex. Interestingly enough, the categorical information in ITC can only be seen at the population level. All in all, it is once again found that the visual stream is organized along a path that goes from being not categorical (early visual areas) to being categorical at the population level while retaining information of within-category individual examples (inferior temporal cortex) to a more fully categorical representation (prefrontal cortex), in agreement with our findings. From the work presented here, we expect an even wider range of such gradient effects of categorical perception to be found experimentally, either with neurophysiology or imagery. 

\section{Conclusion}
Studying categorical perception in biological and artificial neural networks can provide a fruitful discussion between cognitive science and machine learning. Here, we have shown that artificial neural networks that learn a classification task exhibit an enlarged representation near the boundary between categories, with a peak in discrimination, \textit{ie} categorical perception. Our analysis is based on a mathematical understanding and empirical investigations of the geometry of neural representations optimized in view of a classification task, with contraction of space far from category boundaries, and expansion near boundaries. Our results also find counterparts in the literature of neurophysiology and imagery, with low-level generic regions feeding high-level task-specific regions. Our work further suggests a strong gradient of categorical effects along the processing stream, which will be interesting to investigate experimentally.\\

Considering various properties and practical uses of dropout, we have shown that our framework allows to propose a coherent picture of what makes dropout beneficial. We argued that dropout has a differential impact as a function of the neural representation that has been learned so far, implying that its effect is not the same before, during, and after learning, and depends on the layer type (dense vs. convolutional) and depth. Further work is needed so as to quantify more finely how the amount of noise should depend on the level of representation. We believe that our viewpoint should help devising better dropout  protocols, but also new regularization techniques with other types of perturbation than dropout.\\

Another interesting perspective is in the domain of transfer learning, where a neural network trained for a specific task is reused for another task. The measure of categoricality of each layer, which is specific to the classification task at hand, gives thus a measure of the degree (the lack) of genericity of the layer. We expect that this quantity, along with a measure of the overlap between the old and the new tasks, can be used to decide where to cut a neural network for reuse, with the lower part left untouched and the deeper part fine-tuned or retrained from scratch on the new task.\\

To conclude, our work insists on the geometry of internal representations shaped by learning categories, and on the resulting positive impact of noise as a way to learn more robustly. The influence of noise is actually structured by learning, which implies an interaction between these two aspects.

\section*{Materials and Methods}

\paragraph*{Neural distance.}
For a given stimulus $\mathbf{x}$, let us notate $f(\mathbf{x})$ the $N$-dimensional deterministic function computed by the network in the absence of noise (for a given layer with $N$ neurons). The neural distance $D_{\text{neural}}(\mathbf{x}_1, \mathbf{x}_2)$ between two stimuli $\mathbf{x}_1$ and $\mathbf{x}_2$ is then defined, at the population level, as the cosine distance between $f(\mathbf{x}_1)$ and $f(\mathbf{x}_2)$. The cosine distance is equal to 1 minus the cosine similarity, which is equal to the dot product between the two vectors, normalized by the product of the norms of each vector:
\begin{equation}
D_{\text{neural}}(\mathbf{x}_1, \mathbf{x}_2) = 1 - \frac{f(\mathbf{x}_1) \cdot f(\mathbf{x}_2)}{||f(\mathbf{x}_1)||*||f(\mathbf{x}_2)||}
\end{equation}
Note that this measure is not mathematically a distance metric as it does not satisfy the triangular inequality. We nevertheless improperly call this dissimilarity a distance, following a common abuse of language.\\
This cosine distance was chosen so as to easily compare between layers, with different number of neurons: there was a need for a certain normalization, which the cosine distance provides. If one considers normalized vectors, the cosine distance is actually related to the Euclidean distance. Following Fisher information, the neural distance keeps the idea that, as one moves from $x$ to $x+\delta x$, the greater the distance between the activations evoked by these two stimuli the greater the distance in neural space. It would be interesting to consider other measures, that notably take the absolute amplitude of the neural responses into account.

\paragraph*{Categoricality index.}
The categoricality index quantifies the degree of relative intra-class compression vs inter-class expansion of the representation provided by a given layer. It measures the distance between (i) the distribution of the neural distance of items that belong to the same category, and (ii) the distribution of the neural distance of items that are drawn from different categories. Technically, we compute these distributions thanks to random samples taken either from the same categories or from different categories, and we take as distance between the two distributions the two sample Kolmogorov-Smirnov statistic \citep[using the Python implementation provided by the SciPy package,][]{scipy}. Other distance measures could be considered as well, as those previously proposed in the literature. We expect that they would yield overall qualitatively similar results. For instance, \citet{mehrer2020individual} recently considered a clustering index defined as the ``normalized difference in average distances for stimulus pairs from different categories (across) and stimulus pairs from the same category (within): CCI = (across - withing)/(across + within)''. Yet, while quantifying similarly the degree of intra class compression vs inter class separation, we believe our measure gives a better account of the categorical nature of the neural representation by considering not just the average values of the `across' and `within' distances between pairs but the full distributions. Imagine two cases where the average quantities `across' and `within' are equal, but whose distributions exhibit different variances: in the first case, the two distributions of distance are very well separated, with a small variance for each `across' or `within' distribution; in the second case, the two distributions overlap substantially, with larger variance for these distributions. By definition, both cases will receive the same clustering index as defined in \citet{mehrer2020individual}, but our measure assigns a greater categoricality to the first case, as expected by construction of this example.

\paragraph*{Section ``\textbf{\nameref{sec:toy}}'': Estimate of the deterministic counterpart of a noisy neural activity.}
For a given layer, we consider the neural activity  $\mathbf{r}=\{r_1,\ldots,r_N \}$ from a population of $N$ neurons evoked by a stimulus $\mathbf{x}$ as a noisy version $\widetilde{f(\mathbf{x})}$ of the $N$-dimensional deterministic function $f(\mathbf{x})$ computed by the network at that level. As an example in the spirit of the dropout heuristic, for a Gaussian multiplicative noise, $\mathbf{r} = \widetilde{f(\mathbf{x})} = f(\mathbf{x})*\xi$, where $\xi \sim \mathcal{N}(1, \sigma^2)$ ($\sigma^2$ being the noise variance). For a given $\mathbf{x}$ and a given $\mathbf{r} = \widetilde{f(\mathbf{x})}$, we make use of gradient descent to compute the estimate $\widehat{\mathbf{x}}$ that minimizes the square error between $\widetilde{f(\mathbf{x})}$ and $f(\widehat{\mathbf{x}})$:
\begin{equation}
\widehat{\mathbf{x}} \equiv \argmin_{\mathbf{x}^*} \left(\widetilde{f(\mathbf{x})} - f(\mathbf{x}^*)\right)^2.
\end{equation}

\paragraph*{Section ``\textbf{\nameref{sec:mnist_continuum}}'': Autoencoder architecture and learning.}
The autoencoder is a made of an encoder chained with a decoder, both convolutional neural networks. The encoder is made of three convolutional layers, each followed by a max-pooling layer. Similarly, the decoder uses three convolutional layers, each followed by an upsampling layer. All cells have ReLU activation function. The autoencoder is trained with mean square error loss on the full MNIST training set for 1000 epochs, through gradient descent using Adam optimizer \citep{kingma2015adam} with learning rate 1e-4.

\paragraph*{Section ``\textbf{\nameref{sec:mnist_depth}}'': Selection of a set of image continua.} 
We explain here how the continua considered in Section \ref{sec:mnist_depth} are selected. We draw the first 100 samples of each class from the test set. Each $n$th sample from one category is paired with the $n$th sample from another category, leading to 4500 pairs of stimuli drawn from two different digit categories. For each pair we generate a continuum as explained above. Continua presented in Fig.~\ref{fig:mnist_continua} and used in the experiment presented in Section~\ref{sec:mnist_cp}, Fig.~\ref{fig:mnist_cp}, are made of 16 images. Continua used in Section~\ref{sec:mnist_depth}, Fig.~\ref{fig:mnist_cp_depth}, use a finer resolution of 50 images. Not all pairs are valid pairs to look at in the context of our study: we are indeed interested in pairs that can be smoothly interpolated from one category to another. Categories that are close one to another are mainly concerned here -- for instance, if the generated continuum straddles another third category then it should be dismissed from this analysis. In order to only consider the relevant pairs, we keep a pair only if the sum of the two posterior probabilities is above a certain threshold (0.95 here) all along the continuum. In order to average over all these examples, we also exclude cases where the category boundary is be too close to one of the extremities of the continuum. In the end, 1650 pairs fulfill these criteria and are included in the study.

\paragraph*{Section ``\textbf{\nameref{sec:mnist_categoricality}}'': Details on the numerical protocol.} 
The multi-layer perceptron has three hidden layers of 1024 cells, with ReLU activations. Gaussian dropout is used after each dense layer, with respective rate $0.1$, $0.2$, $0.4$. The convolutional neural network has two blocks of two convolutional layers of 32 cells with ReLU activations, followed by a max pooling layer, these two blocks finally followed by a dense hidden layers of $128$ cells. Each block is followed by a dropout layer with rate $0.2$, and the dense layer is followed by a dropout layer with rate $0.5$. Simulations for the multi-layer perceptrons and the convolutional neural networks share the same framework. It considers 10 trials. Each trial uses a different random initialization. For each trial, learning is done over 50 epochs through gradient descent using Adam optimizer with default parameters \citep{kingma2015adam}. In order to compute the categoricality index, the distributions of both the within- and between-category distances are evaluated thanks to 1000 pairs of samples drawn from the same category and 1000 pairs of samples drawn from different categories. Samples come from the test set.

\paragraph*{Section ``\textbf{\nameref{sec:cat_dog}}'': Details on the numerical protocol.} 
We make use of the Kaggle Dogs vs. Cats database\footnote{\url{https://www.microsoft.com/en-us/download/details.aspx?id=54765}, \url{https://www.kaggle.com/c/dogs-vs-cats}}, which contains 25,000 images of dogs and cats. Image size is 180x180. The convolutional neural network has three blocks, each one composed of two convolutional layers of 64 cells with ReLU activations followed by a max pooling layer, these three blocks being finally followed by a global average pooling layer. Such average pooling layer has been introduced in~\citet{lin2013network} to replace the final dense layers of previous common models, reducing drastically the number of parameters -- this avoids overfitting when dealing with not so large databases such as the one considered here. The two first convolutional blocks are followed by a Gaussian dropout layer with rate $0.1$ and $0.2$ respectively, while the global average pooling layer is followed by a dropout layer with rate $0.4$. The simulation is repeated 10 times with different random initializations. Learning is done over 100 epochs through gradient descent using Adam optimizer with default parameters \citep{kingma2015adam}. The learning database being quite small, we used data augmentation during learning, performing horizontal random flip and random rotations (with an angle in $[-0.1 \times 2\pi, 0.1 \times 2\pi]$). After learning, average classification performance on the test set is 95\%. Finally, the different cat/dog, cat/cat or dog/dog morphed continua are generated thanks to the code provided by \citet{miyato2018spectral}\footnote{\url{https://github.com/pfnet-research/sngan_projection}}. We made use of the 256x256 model pretrained on ImageNet that is provided by the authors.

\paragraph*{Section ``\textbf{\nameref{sec:imagenet}}'': Evaluation of the categoricality index for the ImageNet experiment.} 
The full ImageNet database consists in more than a million images categorized into 1000 different classes. Categoricality is evaluated through the use of 1000 pairs of samples for the within-category distribution of neural distances, and 1000 pairs of samples for the between-categories one. Samples come from the validation set. 

\paragraph*{Computer code.} 
The custom Python 3 code written for the present project makes use of the following libraries: \texttt{tensorflow v2.4.1} \citep{tensorflow2015-whitepaper} (using \texttt{tf.keras} API, \citealp{chollet2015keras}), \texttt{matplotlib v3.3.4} \citep{hunter2007matplotlib}, \texttt{numpy v1.19.5} \citep{harris2020array}, \texttt{scipy v1.4.1} \citep{scipy}, \texttt{pandas v1.2.3} \citep{mckinney2010data}, \texttt{seaborn v0.11.1} and \texttt{scikit\_learn v0.24.1} \citep{scikit-learn}. The code is available at \url{https://github.com/l-bg/categorical_perception_ann_neco}.

\section*{Acknowledgments}
We are grateful to Gary Cottrell as well as to an anonymous referee for important and constructive comments. We thank the Information Systems Division (DSI) of the EHESS, and in particular Laurent Henry, for their helpfulness in providing us with an access to computing resources during the covid-19 lockdown.

\clearpage
\renewcommand{\thesection}{Appendix \Alph{section}}
\renewcommand{\thesubsection}{\Alph{section}.\arabic{subsection}}
\setcounter{section}{0} 
\renewcommand{\theequation}{\Alph{section}.\arabic{equation}}
\setcounter{equation}{0} 
\renewcommand{\thefigure}{\Alph{section}.\arabic{figure}} 
\setcounter{figure}{0} 

\section{Modeling categorical perception} 
\label{sec:appendix_model}
For completeness, in this Appendix we review and synthesize the results in \citet{LBG_JPN_2008, LBG_JPN_2012} that are relevant for the present paper. In a companion paper \citep{LBG_JPN_Theory_2020}, we show that the analysis presented here can be extended to multi-layer networks, making explicit that $x$, instead of being the stimulus, corresponds to the projection (achieved by the network) of the stimulus on a space relevant for the discrimination task. 

\subsection{Model Description}
\label{sec:model_description}
We consider a finite set of $M$ categories, denoted by $\mu = 1, \ldots, M$, with  probabilities of occurrence (relative frequency) $q_{\mu} > 0$, so that $\sum_{\mu} q_{\mu}  = 1$. Each category is defined as a probability density distribution $P(\mathbf{x}|\mu)$ over the continuous space of stimulus $\mathbf{x}$. The stimulus space is assumed to be of small dimension $K$, corresponding to the selection of the features or directions relevant to the task at hand. For the sake of simplicity, we only consider here the one-dimensional case, $K=1$. See \citet{LBG_JPN_2008,LBG_JPN_2012}  for equations in the general case $K>1$.\\

A stimulus $x \in \mathbb{R}$ elicits a response $\mathbf{r}=\{r_1,\ldots,r_N \}$ from a population of $N$ neurons. This neural activity $\mathbf{r}$ is some noisy representation of $x$ and aims at encoding properties about the given categories. The read-out is realized by $M$ output cells with activities $g_{\mu}, \mu=1,...,M$. Each output activity is a deterministic function of the neural activity $\mathbf{r}$, $g_{\mu} = g(\mu | \mathbf{r})$. With the goal of computing an estimate $\widehat{\mu}$ of $\mu$, we consider these outputs as estimators of the posterior probability $P(\mu|x)$, where $x$ is the (true) stimulus that elicited the neural activity $\mathbf{r}$. The processing chain can be summarized with the following Markov chain:
\begin{equation}
\mu \rightarrow x \xrightarrow[\text{coding}]{} \mathbf{r} \xrightarrow[\text{decoding}]{} \widehat{\mu} 
\end{equation}

\subsection{Estimation of the posterior probabilities}
The read-out assumes that, given a neural activity $\mathbf{r}$ in the coding layer, the goal is to construct as neural output an estimator $g(\mu|\mathbf{r})$ of the posterior probability $P(\mu|x)$, where $x$ indicates the (true) stimulus that elicited the neural activity $\mathbf{r}$. The relevant Bayesian quality criterion is given by the Kullback-Leibler divergence (or relative entropy) $\mathcal{C}(x,\mathbf{r})$ between the true probabilities $\{P(\mu|x), \mu=1,...,M\}$ and the estimator
$\{g(\mu|\mathbf{r}), \mu=1,...,M\}$, defined as \citep{Cover_Thomas_2006}:
\begin{equation}
\mathcal{C}(x,\mathbf{r}) 
\equiv D_{KL}(P_{\mu|x}||g_{\mu|\mathbf{r}}) 
= \sum_{\mu=1}^{M} P(\mu|x) \ln \frac{P(\mu|x)}{g(\mu|\mathbf{r}) }
\label{eq:cost}
\end{equation}
Averaging over $\mathbf{r}$ given $x$, and then over $x$, the mean cost induced by the estimation can be written:
\begin{equation}
\mathcal{\overline{C}} = - \mathcal{H}(\mu|x) - \int \,\left( \int \,  \sum_\mu P(\mu|x) \ln g(\mu|\mathbf{r})\; P(\mathbf{r}|x)\,d\mathbf{r} \,\right)\; p(x) \,dx 
\label{eq:mean_cost}
\end{equation}
where $\mathcal{H}(\mu|x) = - \int dx\,p(x) \sum_{\mu=1}^{M} P(\mu|x) \ln P(\mu|x)$ is the conditional entropy of $\mu$ given $x$.\\

We can rewrite (\ref{eq:mean_cost}) as the sum of two terms :
\begin{equation}
\mathcal{\overline{C}} = \mathcal{\overline{C}}_{\text{coding}} + \mathcal{\overline{C}}_{\text{decoding}}
\label{eq:mean_cost_coding_decoding}
\end{equation}
respectively defined as : 
\begin{equation}
\mathcal{\overline{C}}_{\text{coding}} = I(\mu,x)- I(\mu,\mathbf{r})
\label{eq:cost_coding}
\end{equation}
and
\begin{equation}
\mathcal{\overline{C}}_{\text{decoding}} = \int \,  D_{KL}(P_{\mu|\mathbf{r}}||g_{\mu|\mathbf{r}}) \;P(\mathbf{r})\,d\mathbf{r}
\label{eq:cost_decoding}
\end{equation}
$I(\mu,x)$ and $I(\mu,\mathbf{r})$ are respectively the mutual information between the categories $\mu$ and the stimulus $x$, and between the categories $\mu$ and the neural activity $\mathbf{r}$, defined by \citep{Blahut_1987}:
\begin{equation}
I(\mu,x) = \sum_{\mu=1}^M q_{\mu}  \int \,\ln \frac{P(x|\mu)}{P(x)} \,  P(x|\mu) \, dx\,, \;\;
I(\mu,\mathbf{r}) = \sum_{\mu=1}^M q_{\mu}  \int \,  \ln \frac{P(\mathbf{r}|\mu)}{P(\mathbf{r})}\; P(\mathbf{r}|\mu) \,d\mathbf{r} 
\label{eq:mi_mu_x}
\end{equation}
$D_{KL}(P_{\mu|\mathbf{r}}||g_{\mu|\mathbf{r}})$  is the relative entropy between the true probability of the category given the neural activity and the output function $g$:
\begin{equation}
D_{KL}(P_{\mu|\mathbf{r}}||g_{\mu|\mathbf{r}}) = \sum_{\mu=1}^{M} P(\mu|\mathbf{r}) \ln \frac{P(\mu|\mathbf{r})}{g(\mu|\mathbf{r}) }
\label{eq:dklpgr}
\end{equation}

Since processing cannot increase information \citep[see e.g.][pp. 158-159]{Blahut_1987}, the information $I(\mu,\mathbf{r})$ conveyed by $\mathbf{r}$ about $\mu$ is at most equal to the one conveyed by the sensory input $x$, hence we have that $\mathcal{\overline{C}}_{\text{coding}} \geq 0$. This coding cost tends to zero as noise vanishes. The decoding cost $\mathcal{\overline{C}}_{\text{decoding}}$ is the only term that depends on $g$, hence the function minimizing the cost function (\ref{eq:mean_cost_coding_decoding}) is (if it can be realized by the network):
\begin{equation}
g(\mu|\mathbf{r}) = P(\mu|\mathbf{r})
\label{eq:optdecod}
\end{equation}

From a machine learning viewpoint, minimization of the decoding cost (\ref{eq:dklpgr}) can be achieved through supervised learning taking as cost function the cross-entropy loss, as shown in \citet{LBG_JPN_2012}, SI Section 1.

\subsection{Coding efficiency}
\label{sec:coding_efficiency}

In \citet{LBG_JPN_2008}, we show that, in a high signal-to-noise ratio limit, that is when the number $N$ of coding cells grows to infinity, the mutual information $I(\mu,\mathbf{r})$ between the activity of the neural population and the set of discrete categories reaches its upper bound, which is the mutual information $I(\mu,x)$ between the stimuli and the categories. For large but finite $N$, the leading correction takes an interesting form. It can be written as the average (over the stimulus space) of the ratio between two Fisher-information values: in the denominator, the Fisher information $ F_{\text{code}}(x)$, specific to the neural encoding stage $x \rightarrow \mathbf{r}$, and in the numerator, the Fisher information $ F_{\text{cat}}(x)$, that characterizes the category realizations $\mu \rightarrow x$ and does not depend on the neural code. Fisher information is an important concept that comes from the field of parameter estimation in statistics. $F_{\text{code}}(x)$ characterizes the sensitivity of the neural activity $\mathbf{r}$ with respect to small variations of $x$. The higher the Fisher information $F_{\text{code}}(x)$, the better an estimate of $x$ can be obtained. $F_{\text{cat}}(x)$ quantifies the categorization uncertainty. As a consequence, $F_{\text{cat}}(x)$ is larger in the transition regions between categories, where the identification function $P(\mu|x)$ changes quickly, than within category, where the identification function $P(\mu|x)$ is almost flat.\\

Explicitly, the coding cost~(\ref{eq:cost_coding})  writes: 
\begin{equation}
\mathcal{\overline{C}}_{\text{coding}} = \frac{1}{2} \int  \frac{F_{\text{cat}}(x)}{F_{\text{code}}(x)}\;p(x)\,dx
\label{eq:midiff_app}
\end{equation}
where $F_{\text{code}}(x)$ and $F_{\text{cat}}(x)$ are respectively defined as
\begin{equation}
F_{\text{code}}(x) = - \int \,  \frac{\partial^2 \ln P(\mathbf{r}|x) }{\partial x^2} \;P(\mathbf{r}|x) \,d\mathbf{r} 
\label{eq:fisher_code}
\end{equation}
\begin{equation}
F_{\text{cat}}(x) = -\sum_{\mu=1}^M \,  \frac{\partial^2 \ln P(\mu|x)}{\partial x^2} \; P(\mu|x). \,\\
\label{eq:fisher_cat}
\end{equation}
Crucially for the present work, the inverse of the Fisher information is an optimal lower bound on the variance $\sigma_x^2 $ of any unbiased estimator $\widehat{x}(\mathbf{r})$ of $x$ \citep[Cramér-Rao bound, see e.g.][]{Blahut_1987}:
\begin{equation}
\sigma_x^2 \equiv \int \,  \big(\widehat{x}(\mathbf{r}) - x\big)^2 \;P(\mathbf{r}|x)\,d\mathbf{r}\;\geq \; \frac{1}{F_{\text{code}}(x)}
\label{eq:cramer_rao_app}
\end{equation}

In the case of $x$ in dimension $K>1$, one has $K\times K$ Fisher information matrices $F_{\text{cat}}$ and $F_{\text{code}}$, the Cramér-Rao bound relates the covariance matrix of the estimator to the inverse of $F_{\text{code}}$, and the ratio of the Fisher information quantities in (\ref{eq:midiff_app}) is replaced by the trace of the product of the transpose of $F_{\text{cat}}$ by the inverse of $F_{\text{code}}$ \citep[see][]{LBG_JPN_2008}. In such case the resulting neural metric will be anisotropic. Suppose one follows a 1d path (a continuum between two items). If this path crosses a category boundary, one will observe the same contraction/expansion of space as for the 1d case, whereas along a path parallel to a boundary the neural Fisher information will be more or less constant.

\subsection{Optimal decoding}
\label{sec:opt_decoding}

In \citet{LBG_JPN_2012}, we show that the function $g(\mu|\mathbf{r})$ that minimizes the decoding cost function given by Eq. \ref{eq:cost_decoding} is equal to $P(\mu|\mathbf{r})$, which is an (asymptotically) unbiased and (asymptotically) efficient estimator of $P(\mu|x)$. For a given $x$, its mean is thus equal to
\begin{equation}
\int  \, g(\mu|\mathbf{r}) \, P(\mathbf{r}|x)\,d\mathbf{r} = P(\mu|x)
\end{equation}
and its variance is given by the Cramér-Rao bound, that is, in this 1d case, 
\begin{equation}
\int \, \big(g(\mu|\mathbf{r})-P(\mu|x)\big)^2 \;P(\mathbf{r}|x) \, d\mathbf{r}\;
=\; \frac{P'(\mu|x)^2}{F_\text{code}(x)}
\label{eq:pmur_cramer_rao}
\end{equation}

\subsection{Category learning implies categorical perception}
\label{sec:learning-cp}
The Fisher information $F_{\text{cat}}(x)$ is the largest at the boundary between categories. If the neural code is to be optimized, from Eq.~\ref{eq:midiff_app} we therefore expect the Fisher information $F_{\text{code}}(x)$ to be greater between categories than within, so as to compensate for the higher value of $F_{\text{cat}}(x)$ in this region. Depending on the constraints specific to the system under consideration, minimization of the cost leads to a neural code such that $F_{\text{code}}(x)$ is some increasing function of $F_{\text{cat}}(x)$. For some constraints one gets $F_{\text{code}}(x)\propto F_{\text{cat}}(x)$ as optimum, but other constraints may lead to other relationships -- see \citet{LBG_JPN_2008,LBG_JPN_Theory_2020, KB_JPN_2020}. \\

\begin{figure}[hb]
	\centering
	\includegraphics[width=0.98\linewidth]{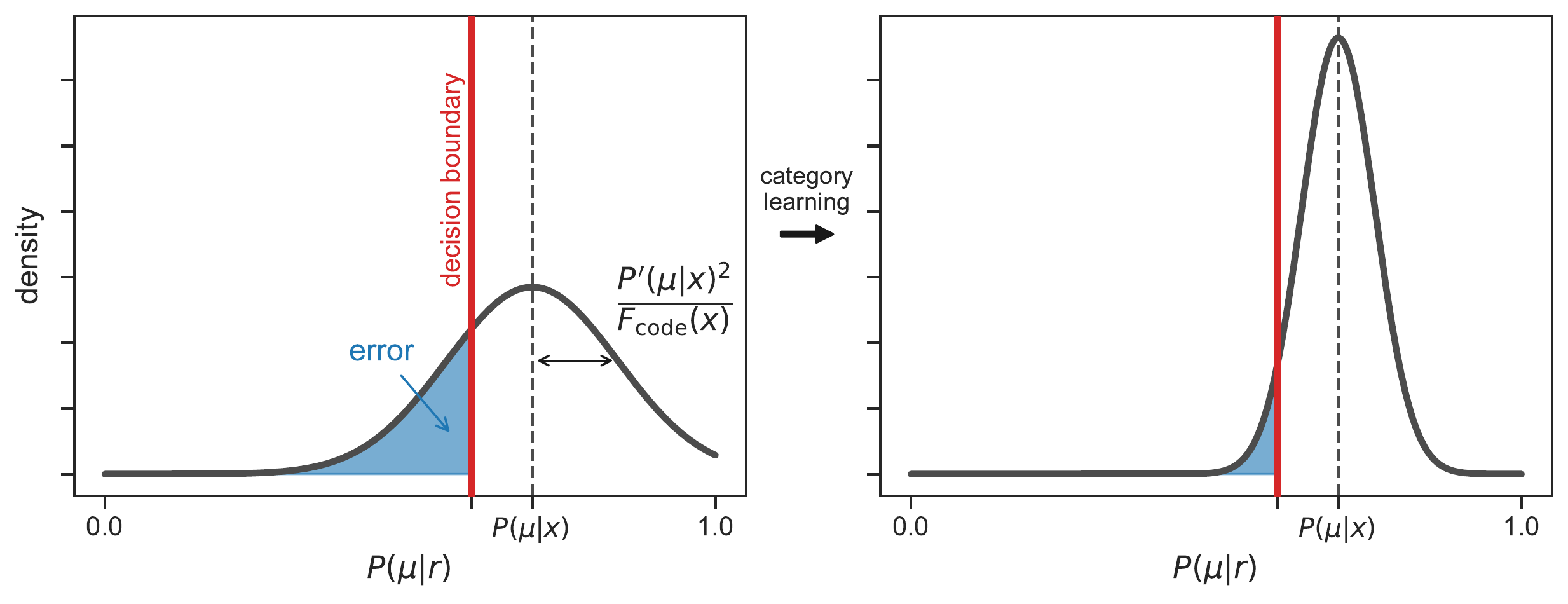}
	\caption{\textbf{Increasing Fisher information $F_\text{code}$ decreases the probability of error due to noise in the neural processing.}
		Probability of misclassifying a stimulus $x$ given the neural activity $\mathbf{r}$ that it has evoked is greater the closer to the boundary between categories. Here, we consider an example $x$ that belongs to category $\mu$. In order to minimize the probability of error, the best strategy is to categorize $x$ as $\mu$ if $\mu$ has the greatest posterior probability $P(\mu|\mathbf{r})$, above some value defining the decision boundary. All the values of  $P(\mu|\mathbf{r})$ below the decision boundary result in an error in classification. Learning the categories increases the Fisher information $F_\text{code}$ at the boundary between categories, which reduces the variance $P'(\mu|x)^2/F_{\text{code}}(x)$ of the estimator of $P(\mu|x)$, thus reducing the probability of error.}
	\label{fig:posterior_estimation}
\end{figure}

Another way to look at the benefit of having greater neural sensitivity in the transition region between categories is through Eq.~\ref{eq:pmur_cramer_rao}: a greater Fisher information $F_{\text{code}}(x)$ in this region, where $P'(\mu|x)^2$ is the highest, makes it possible to lower the variance of the estimate $g(\mu|\mathbf{r})$ of the $P(\mu|x)$. As a result, the probability of misclassifying $x$ given the neural activity $\mathbf{r}$ is also decreased, as illustrated in Fig.~\ref{fig:posterior_estimation}. \\

Larger Fisher information $F_{\text{code}}(x)$ means greater sensitivity of the neural code to a small change in stimulus $x$. $F_{\text{code}}$ gives the metric of the representation: a larger value around a certain $x$ means that the representation is dilated at that location. In other words, category learning implies better cross-category than within-category discrimination, hence the so-called \textit{categorical perception}. 

\clearpage
\setcounter{figure}{0} 
\section{Examples of image continua, using the MNIST dataset}
\label{sec:appendix_mnist_continua_examples}
$\,$\\

\begin{figure}[h]
	\centering
	\includegraphics[width=0.95\linewidth]{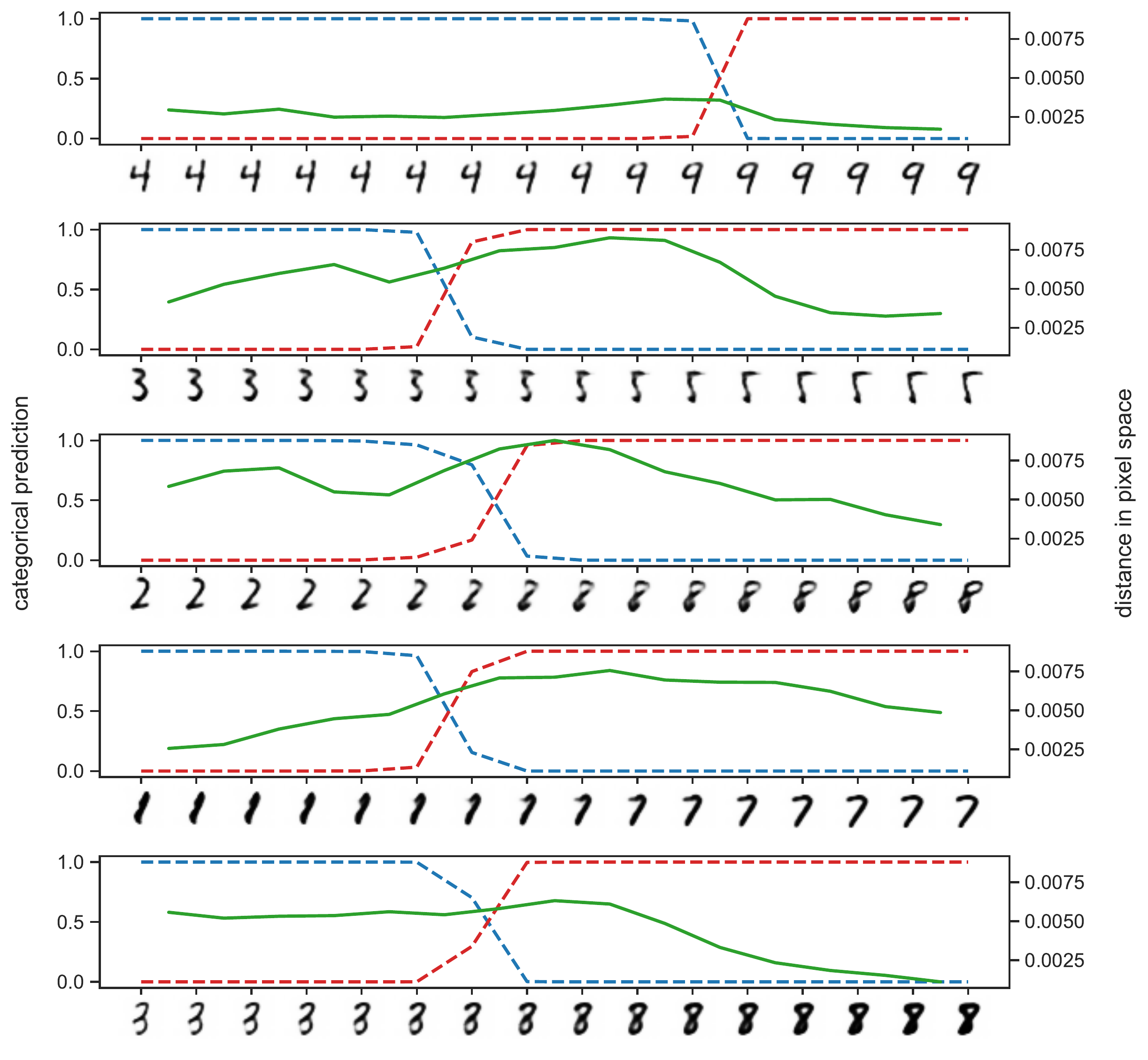}
	\caption{\textbf{Examples of image continua}, smoothly interpolating between pairs of digits taken from the MNIST test set. The dotted colored lines indicate the posterior probabilities from the network of Fig.~\ref{fig:mnist_cp_depth} (blue corresponds to the correct response for the leftmost digit, red for the rightmost one).
	The green solid line corresponds to the distance in input (pixel) space between adjacent items along the continuum.}
	\label{fig:mnist_continua_examples_input_distance_pred}
\end{figure}

\clearpage
\setcounter{figure}{0} 
\section{Supplementary figure for the cat/dog example: Distance in input space} 
\label{sec:appendix_catdog_pixel}
$\,$\\
\begin{figure}[h]
	\textbf{a.}
	\includegraphics[width=.95\linewidth,valign=T]{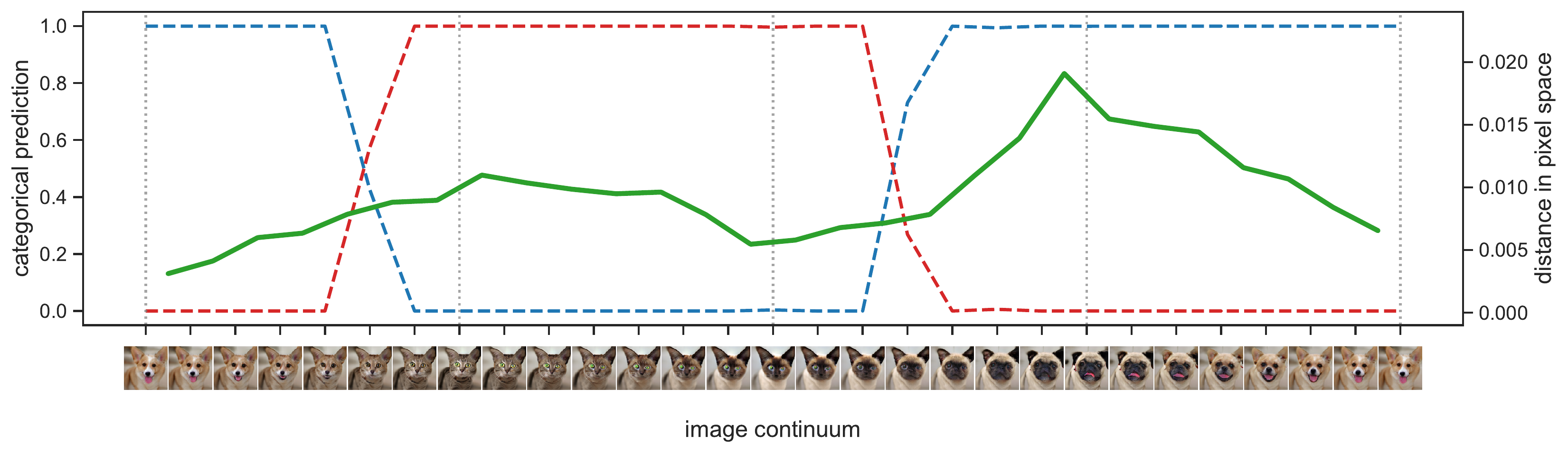}
	\vspace{0.1cm}\\
	\textbf{b.}
	\includegraphics[width=.95\linewidth,valign=T]{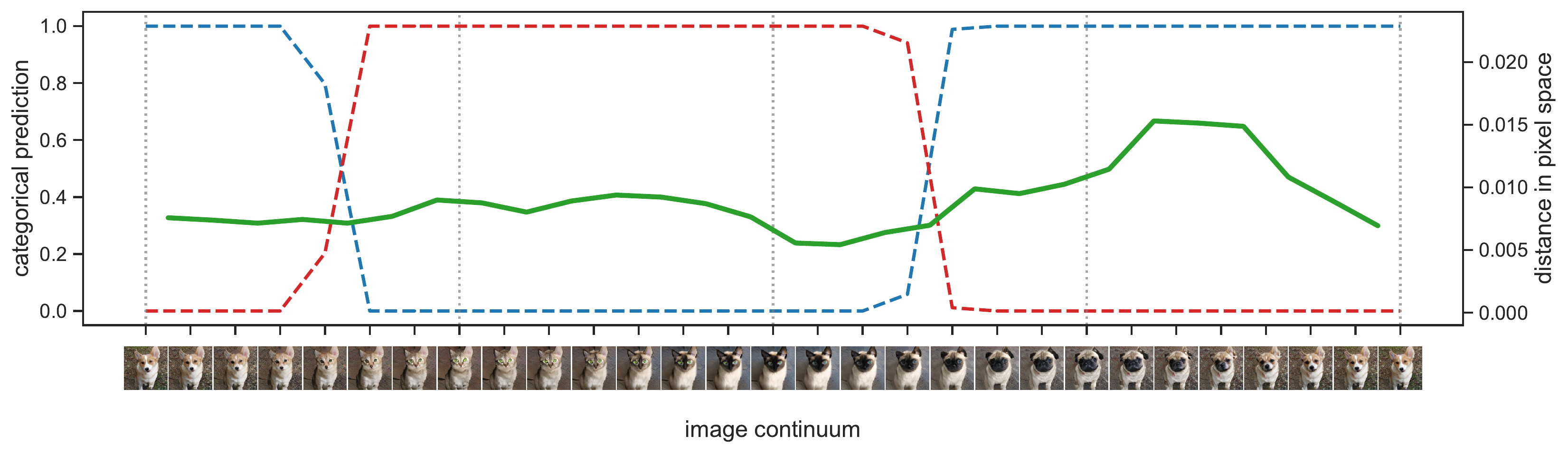}
	\caption{\textbf{Categorical perception of a cat/dog continuum: Distance in input space}. Same legend as in Fig.~\ref{fig:cat_dog}. The green solid line corresponds to the distance in input (pixel) space between adjacent items along the continuum.}
	\label{fig:cat_dog_input}
\end{figure}

\clearpage
\setcounter{figure}{0} 
\section{Supplementary results on dropout} 
\label{sec:appendix_dropout}

\subsection{Dropout rate vs. layer depth}
\label{sec:appendix_dropout_depth}

We propose here a first investigation aiming at testing the effectiveness of dropout as a function of layer depth. We conduct an experiment using a network trained on the CIFAR-10 image dataset \citep{krizhevsky2009learning}. This dataset is a collection of 60000 images (with 50000 images in the training set and 10000 in the test set) divided into 10 classes (such as airplanes, cars, birds or cats). The neural network is a multi-layer perceptron with two hidden layers of 1024 cells, with ReLU activations. We apply dropout after the input layer and after each hidden layer. For each of these three cases, we vary the dropout rate from 0.0 to 0.7 while keeping the other two with a fixed dropout rate of 0.2. For each value of dropout, the experiment consists in 10 trials with different random initializations. For each trial, learning is done over 1000 epochs through gradient descent using Adam optimizer with learning rate 1e-4 \citep{kingma2015adam}. In order to accommodate for the fact that different values of the dropout rate might require different learning rates or numbers of epochs \citep{srivastava2014dropout}, we consider the average of the ten best values obtained on the test set during these 1000 epochs (instead of considering only the very last value).\\

\begin{figure}[h]
	\centering
	\includegraphics[width=0.6\linewidth]{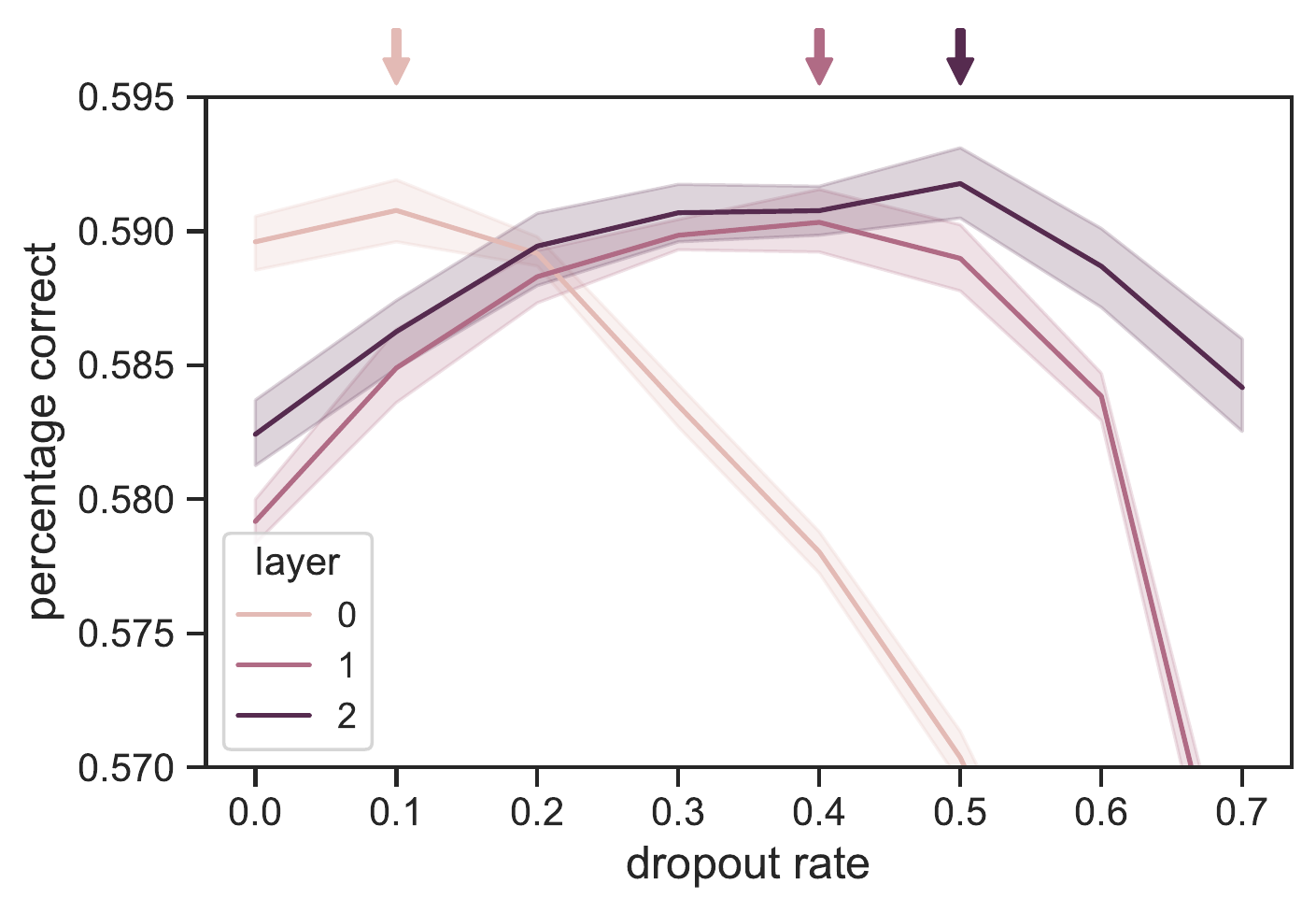}
	\caption{\textbf{Classification accuracy on the CIFAR-10 image dataset (test set) using a multi-layer perceptron with varying levels of dropout.} The neural network is a multi-layer perceptron with two hidden layers. Dropout is applied after the input layer (called layer 0 here) and the two hidden layers (layer 1 and 2). 
	For a given layer, we vary the dropout rate from 0.0 to 0.7 while keeping a fixed rate value of 0.2 in the other two layers. 
	Each arrow marks the maximum value in classification accuracy for the corresponding layer. Error bars indicate 95\% bootstrap confidence intervals.}
	\label{fig:cifar10_dropout}
\end{figure}

Fig.~\ref{fig:cifar10_dropout} presents the results. First, one can see that the deeper the layer the more robust it is to a great amount of noise: while a large amount of dropout rate (see e.g. 0.7) is very detrimental when applied at the level of the input layer, the deepest hidden layer, which is more categorical, does not suffer much from it. Second, one can see that the best dropout value for each layer increases with depth, which supports the idea that a deep layer benefits more from a larger value of dropout.

\clearpage
\subsection{Comparing categoricality on the MNIST dataset with and without dropout}
\label{sec:appendix_dropout_mnist}

\begin{figure}[h]
	\centering
	\textbf{a.}
	\includegraphics[width=0.44\linewidth,valign=T]{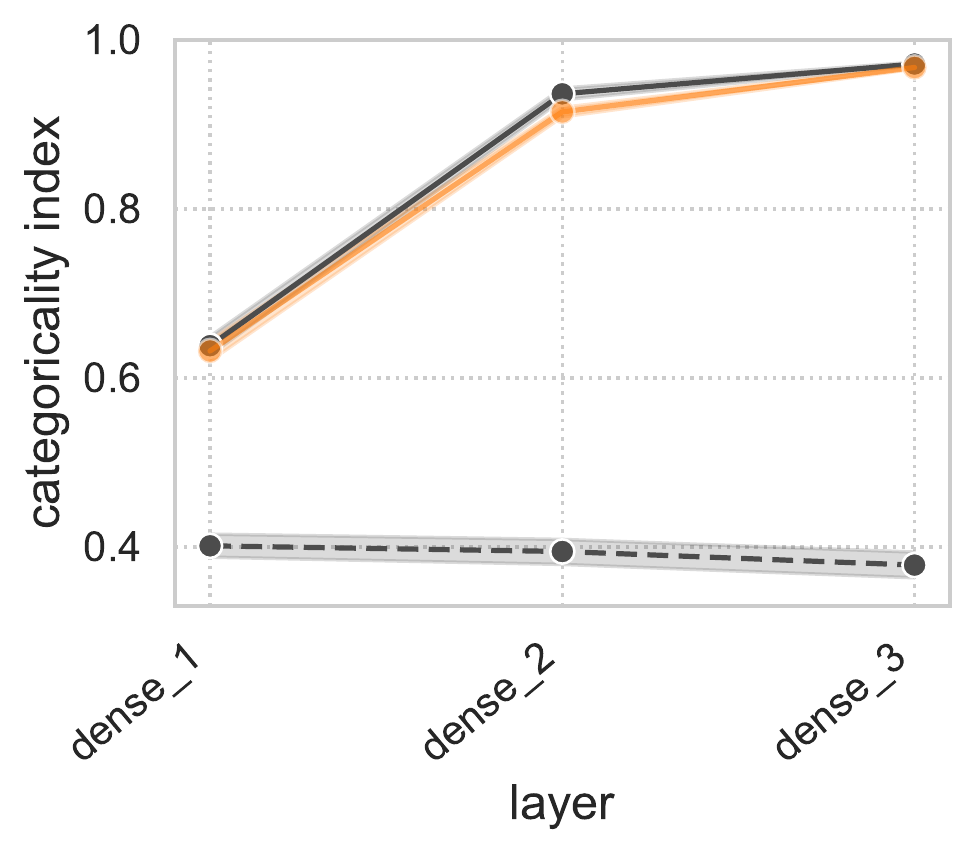}
	\hfill
	\textbf{b.}
	\includegraphics[width=0.44\linewidth,valign=T]{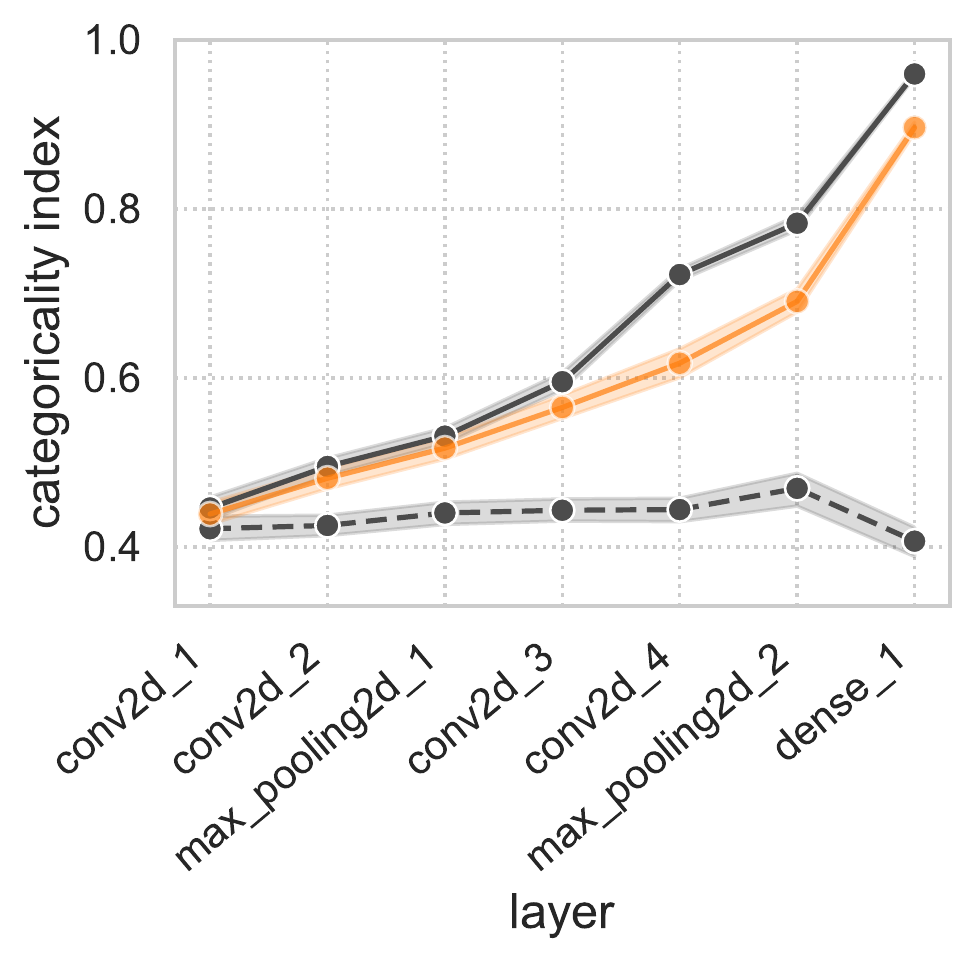}
    \caption{\textbf{Categoricality as a function of layer depth, using the MNIST dataset, with and without the use of dropout}. Reproduction of Fig.~\ref{fig:mnist_categoricality}, along with the categoricality obtained without the use of dropout or Gaussian dropout (orange lines).}
	\label{fig:mnist_categoricality_nodropout}
\end{figure}

\clearpage

\bibliographystyle{apalike}
\bibliography{refs_cpdl.bib}

\begin{thebibliography}{}

\bibitem[Abadi et~al., 2015]{tensorflow2015-whitepaper}
Abadi, M., Agarwal, A., Barham, P., Brevdo, E., Chen, Z., Citro, C., Corrado,
  G.~S., Davis, A., Dean, J., Devin, M., Ghemawat, S., Goodfellow, I., Harp,
  A., Irving, G., Isard, M., Jia, Y., Jozefowicz, R., Kaiser, L., Kudlur, M.,
  Levenberg, J., Man\'{e}, D., Monga, R., Moore, S., Murray, D., Olah, C.,
  Schuster, M., Shlens, J., Steiner, B., Sutskever, I., Talwar, K., Tucker, P.,
  Vanhoucke, V., Vasudevan, V., Vi\'{e}gas, F., Vinyals, O., Warden, P.,
  Wattenberg, M., Wicke, M., Yu, Y., and Zheng, X. (2015).
\newblock {TensorFlow}: Large-scale machine learning on heterogeneous systems.
\newblock Software available from tensorflow.org.

\bibitem[Abramson and Lisker, 1970]{abramson1970discriminability}
Abramson, A.~S. and Lisker, L. (1970).
\newblock Discriminability along the voicing continuum: Cross-language tests.
\newblock In {\em Proceedings of the sixth international congress of phonetic
  sciences}, volume 196, pages 569--573. Prague.

\bibitem[Alain and Bengio, 2016]{alain2016understanding}
Alain, G. and Bengio, Y. (2016).
\newblock Understanding intermediate layers using linear classifier probes.
\newblock {\em arXiv preprint arXiv:1610.01644}.

\bibitem[An, 1996]{an1996effects}
An, G. (1996).
\newblock The effects of adding noise during backpropagation training on a
  generalization performance.
\newblock {\em Neural computation}, 8(3):643--674.

\bibitem[Anderson et~al., 1977]{anderson1977distinctive}
Anderson, J.~A., Silverstein, J.~W., Ritz, S.~A., and Jones, R.~S. (1977).
\newblock Distinctive features, categorical perception, and probability
  learning: Some applications of a neural model.
\newblock {\em Psychological review}, 84(5):413.

\bibitem[Beale and Keil, 1995]{Beale_Keil_1995}
Beale, J. and Keil, F. (1995).
\newblock Categorical effects in the perception of faces.
\newblock {\em Cognition}, 57:217--239.

\bibitem[Bengio et~al., 2009]{bengio2009curriculum}
Bengio, Y., Louradour, J., Collobert, R., and Weston, J. (2009).
\newblock Curriculum learning.
\newblock In {\em Proceedings of the 26th annual international conference on
  machine learning}, pages 41--48.

\bibitem[Bengio et~al., 2013]{bengio2013better}
Bengio, Y., Mesnil, G., Dauphin, Y., and Rifai, S. (2013).
\newblock Better mixing via deep representations.
\newblock In {\em International conference on machine learning}, pages
  552--560.

\bibitem[Berlemont and Nadal, 2020]{KB_JPN_2020}
Berlemont, K. and Nadal, J.-P. (2020).
\newblock Confidence-controlled hebbian learning efficiently extracts category
  membership from stimuli encoded in view of a categorization task.
\newblock {\em bioRxiv preprint, to appear in Neural Computation}.
\newblock \url{https://doi.org/10.1101/2020.08.06.239533}. To appear in Neural
  Computation.

\bibitem[Bidelman et~al., 2013]{bidelman2013tracing}
Bidelman, G.~M., Moreno, S., and Alain, C. (2013).
\newblock Tracing the emergence of categorical speech perception in the human
  auditory system.
\newblock {\em Neuroimage}, 79:201--212.

\bibitem[Bishop, 1995]{bishop1995training}
Bishop, C.~M. (1995).
\newblock Training with noise is equivalent to tikhonov regularization.
\newblock {\em Neural computation}, 7(1):108--116.

\bibitem[Blahut, 1987]{Blahut_1987}
Blahut, R.~E. (1987).
\newblock {\em Principles and practice of information theory}.
\newblock Addison-Wesley Longman Publishing Co., Inc., Boston, MA, USA.

\bibitem[Bonnasse-Gahot and Nadal, 2008]{LBG_JPN_2008}
Bonnasse-Gahot, L. and Nadal, J.-P. (2008).
\newblock Neural coding of categories: Information efficiency and optimal
  population codes.
\newblock {\em Journal of Computational Neuroscience}, 25(1):169--87.

\bibitem[Bonnasse-Gahot and Nadal, 2012]{LBG_JPN_2012}
Bonnasse-Gahot, L. and Nadal, J.-P. (2012).
\newblock Perception of categories: from coding efficiency to reaction times.
\newblock {\em Brain Research}, 1434:47--61.

\bibitem[Bonnasse-Gahot and Nadal, 2020]{LBG_JPN_Theory_2020}
Bonnasse-Gahot, L. and Nadal, J.-P. (2020).
\newblock Category learning in deep neural networks: Information content and
  geometry of internal representations.
\newblock {\em In preparation}.

\bibitem[Bornstein and Korda, 1984]{Bornstein_Korda_1984}
Bornstein, M. and Korda, N. (1984).
\newblock Discrimination and matching within and between hues measured by
  reaction times: some implications for categorical perception and levels of
  information processing.
\newblock {\em Psychological Research}, 46:207--222.

\bibitem[Bouthillier et~al., 2015]{bouthillier2015dropout}
Bouthillier, X., Konda, K., Vincent, P., and Memisevic, R. (2015).
\newblock Dropout as data augmentation.
\newblock {\em arXiv preprint arXiv:1506.08700}.

\bibitem[Burns and Ward, 1978]{burns1978categorical}
Burns, E.~M. and Ward, W.~D. (1978).
\newblock Categorical perception—phenomenon or epiphenomenon: Evidence from
  experiments in the perception of melodic musical intervals.
\newblock {\em The Journal of the Acoustical Society of America},
  63(2):456--468.

\bibitem[Caves et~al., 2018]{caves2018categorical}
Caves, E.~M., Green, P.~A., Zipple, M.~N., Peters, S., Johnsen, S., and
  Nowicki, S. (2018).
\newblock Categorical perception of colour signals in a songbird.
\newblock {\em Nature}, 560(7718):365--367.

\bibitem[Chang et~al., 2010]{chang2010categorical}
Chang, E.~F., Rieger, J.~W., Johnson, K., Berger, M.~S., Barbaro, N.~M., and
  Knight, R.~T. (2010).
\newblock Categorical speech representation in human superior temporal gyrus.
\newblock {\em Nature neuroscience}, 13(11):1428.

\bibitem[Chollet et~al., 2015]{chollet2015keras}
Chollet, F. et~al. (2015).
\newblock Keras.
\newblock \url{https://keras.io}.

\bibitem[Cover and Thomas, 2006]{Cover_Thomas_2006}
Cover, T. and Thomas, J. (2006).
\newblock {\em Elements of Information Theory}.
\newblock Wiley \& Sons, NY, USA.
\newblock Second Edition.

\bibitem[Cross et~al., 1965]{cross1965identification}
Cross, D., Lane, H., and Sheppard, W. (1965).
\newblock Identification and discrimination functions for a visual continuum
  and their relation to the motor theory of speech perception.
\newblock {\em Journal of Experimental Psychology}, 70(1):63.

\bibitem[Damper and Harnad, 2000]{Damper_Harnad_2000}
Damper, R. and Harnad, S. (2000).
\newblock Neural network models of categorical perception.
\newblock {\em Percept. Psychophys.}, 62(4):843--867.

\bibitem[Deng et~al., 2009]{deng2009imagenet}
Deng, J., Dong, W., Socher, R., Li, L.-J., Li, K., and Fei-Fei, L. (2009).
\newblock Imagenet: A large-scale hierarchical image database.
\newblock In {\em 2009 IEEE conference on computer vision and pattern
  recognition}, pages 248--255.

\bibitem[Freedman et~al., 2001]{freedman2001categorical}
Freedman, D.~J., Riesenhuber, M., Poggio, T., and Miller, E.~K. (2001).
\newblock Categorical representation of visual stimuli in the primate
  prefrontal cortex.
\newblock {\em Science}, 291(5502):312--316.

\bibitem[Freedman et~al., 2003]{freedman2003comparison}
Freedman, D.~J., Riesenhuber, M., Poggio, T., and Miller, E.~K. (2003).
\newblock A comparison of primate prefrontal and inferior temporal cortices
  during visual categorization.
\newblock {\em Journal of Neuroscience}, 23(12):5235--5246.

\bibitem[Goldstone, 1994]{Goldstone_1994}
Goldstone, R. (1994).
\newblock Influences of categorization on perceptual discrimination.
\newblock {\em Journal of Experimental Psychology: General}, 123(2):178--200.

\bibitem[Goodfellow et~al., 2016]{Goodfellow-et-al-2016}
Goodfellow, I., Bengio, Y., and Courville, A. (2016).
\newblock {\em Deep Learning}.
\newblock MIT Press.
\newblock \url{http://www.deeplearningbook.org}.

\bibitem[Goodfellow et~al., 2014]{goodfellow2014generative}
Goodfellow, I., Pouget-Abadie, J., Mirza, M., Xu, B., Warde-Farley, D., Ozair,
  S., Courville, A., and Bengio, Y. (2014).
\newblock Generative adversarial nets.
\newblock In {\em Advances in neural information processing systems}, pages
  2672--2680.

\bibitem[Goto, 1971]{goto1971auditory}
Goto, H. (1971).
\newblock Auditory perception by normal japanese adults of the sounds ``l'' and
  ``r.''.
\newblock {\em Neuropsychologia}, 9(3):317--323.

\bibitem[Green et~al., 1966]{green1966signal}
Green, D.~M., Swets, J.~A., et~al. (1966).
\newblock {\em Signal detection theory and psychophysics}, volume~1.
\newblock Wiley New York.

\bibitem[Harnad, 1987]{Harnad_1987}
Harnad, S., editor (1987).
\newblock {\em Categorical Perception: The Groundwork of Cognition}.
\newblock New York: Cambridge University Press.

\bibitem[Harris et~al., 2020]{harris2020array}
Harris, C.~R., Millman, K.~J., van~der Walt, S.~J., Gommers, R., Virtanen, P.,
  Cournapeau, D., Wieser, E., Taylor, J., Berg, S., Smith, N.~J., et~al.
  (2020).
\newblock Array programming with numpy.
\newblock {\em Nature}, 585(7825):357--362.

\bibitem[Holmstrom and Koistinen, 1992]{holmstrom1992using}
Holmstrom, L. and Koistinen, P. (1992).
\newblock Using additive noise in back-propagation training.
\newblock {\em IEEE transactions on neural networks}, 3(1):24--38.

\bibitem[Hunter, 2007]{hunter2007matplotlib}
Hunter, J.~D. (2007).
\newblock Matplotlib: A 2d graphics environment.
\newblock {\em Computing in Science \& Engineering}, 9(3):90--95.

\bibitem[Iverson and Kuhl, 2000]{iverson2000perceptual}
Iverson, P. and Kuhl, P.~K. (2000).
\newblock Perceptual magnet and phoneme boundary effects in speech perception:
  Do they arise from a common mechanism?
\newblock {\em Perception \& psychophysics}, 62(4):874--886.

\bibitem[Kingma and Ba, 2015]{kingma2015adam}
Kingma, D.~P. and Ba, J. (2015).
\newblock Adam: A method for stochastic optimization.
\newblock In {\em Proceedings of the 3rd International Conference on Learning
  Representations (ICLR)}.

\bibitem[Kluender et~al., 1998]{Kluender_etal_1998}
Kluender, K., Lotto, A., Holt, L., and Bloedel, S. (1998).
\newblock Role of experience for language-specific functional mappings of vowel
  sounds.
\newblock {\em Journal of the Acoustical Society of America},
  104(6):3568--3582.

\bibitem[Kreiman et~al., 2000]{kreiman2000category}
Kreiman, G., Koch, C., and Fried, I. (2000).
\newblock Category-specific visual responses of single neurons in the human
  medial temporal lobe.
\newblock {\em Nature neuroscience}, 3(9):946--953.

\bibitem[Kriegeskorte et~al., 2008]{kriegeskorte2008matching}
Kriegeskorte, N., Mur, M., Ruff, D.~A., Kiani, R., Bodurka, J., Esteky, H.,
  Tanaka, K., and Bandettini, P.~A. (2008).
\newblock Matching categorical object representations in inferior temporal
  cortex of man and monkey.
\newblock {\em Neuron}, 60(6):1126--1141.

\bibitem[Krizhevsky, 2009]{krizhevsky2009learning}
Krizhevsky, A. (2009).
\newblock Learning multiple layers of features from tiny images.
\newblock {\em Master's thesis, Department of Computer Science, University of
  Toronto}.

\bibitem[Krizhevsky et~al., 2012]{krizhevsky2012imagenet}
Krizhevsky, A., Sutskever, I., and Hinton, G.~E. (2012).
\newblock Imagenet classification with deep convolutional neural networks.
\newblock In {\em Advances in neural information processing systems}, pages
  1097--1105.

\bibitem[Lane, 1965]{lane1965motor}
Lane, H. (1965).
\newblock The motor theory of speech perception: A critical review.
\newblock {\em Psychological Review}, 72(4):275.

\bibitem[LeCun et~al., 2015]{lecun2015deep}
LeCun, Y., Bengio, Y., and Hinton, G. (2015).
\newblock Deep learning.
\newblock {\em Nature}, 521(7553):436--444.

\bibitem[LeCun et~al., 1998]{lecun1998gradient}
LeCun, Y., Bottou, L., Bengio, Y., and Haffner, P. (1998).
\newblock Gradient-based learning applied to document recognition.
\newblock {\em Proceedings of the IEEE}, 86(11):2278--2324.

\bibitem[Liberman et~al., 1957]{Liberman_etal_1957}
Liberman, A., Harris, K., Hoffman, H., and Griffith, B. (1957).
\newblock The discrimination of speech sounds within and across phoneme
  boundaries.
\newblock {\em Journal of Experimental Psychology}, 54:358--369.

\bibitem[Liberman et~al., 1961]{liberman1961effect}
Liberman, A., Harris, K.~S., Eimas, P., Lisker, L., and Bastian, J. (1961).
\newblock An effect of learning on speech perception: The discrimination of
  durations of silence with and without phonemic significance.
\newblock {\em Language and Speech}, 4(4):175--195.

\bibitem[Lin et~al., 2013]{lin2013network}
Lin, M., Chen, Q., and Yan, S. (2013).
\newblock Network in network.
\newblock {\em arXiv preprint arXiv:1312.4400}.

\bibitem[Matsuoka, 1992]{matsuoka1992noise}
Matsuoka, K. (1992).
\newblock Noise injection into inputs in back-propagation learning.
\newblock {\em IEEE Transactions on Systems, Man, and Cybernetics},
  22(3):436--440.

\bibitem[McClelland and Rogers, 2003]{mcclelland2003parallel}
McClelland, J.~L. and Rogers, T.~T. (2003).
\newblock The parallel distributed processing approach to semantic cognition.
\newblock {\em Nature reviews neuroscience}, 4(4):310--322.

\bibitem[McKinney et~al., 2010]{mckinney2010data}
McKinney, W. et~al. (2010).
\newblock Data structures for statistical computing in python.
\newblock In {\em Proceedings of the 9th Python in Science Conference}, volume
  445, pages 51--56. Austin, TX.

\bibitem[Mehrer et~al., 2020]{mehrer2020individual}
Mehrer, J., Spoerer, C.~J., Kriegeskorte, N., and Kietzmann, T.~C. (2020).
\newblock Individual differences among deep neural network models.
\newblock {\em Nature Communications}, 11(1):5725.

\bibitem[Meyers et~al., 2008]{meyers2008dynamic}
Meyers, E.~M., Freedman, D.~J., Kreiman, G., Miller, E.~K., and Poggio, T.
  (2008).
\newblock Dynamic population coding of category information in inferior
  temporal and prefrontal cortex.
\newblock {\em Journal of neurophysiology}, 100(3):1407--1419.

\bibitem[Miyato et~al., 2018]{miyato2018spectral}
Miyato, T., Kataoka, T., Koyama, M., and Yoshida, Y. (2018).
\newblock Spectral normalization for generative adversarial networks.
\newblock In {\em International Conference on Learning Representations}.

\bibitem[Morerio et~al., 2017]{morerio2017curriculum}
Morerio, P., Cavazza, J., Volpi, R., Vidal, R., and Murino, V. (2017).
\newblock Curriculum dropout.
\newblock In {\em Proceedings of the IEEE International Conference on Computer
  Vision}, pages 3544--3552.

\bibitem[Nadal and Parga, 1994]{nadal1994nonlinear}
Nadal, J.-P. and Parga, N. (1994).
\newblock Nonlinear neurons in the low-noise limit: a factorial code maximizes
  information transfer.
\newblock {\em Network: Computation in Neural Systems}, 5(4):565--581.

\bibitem[Nelson and Marler, 1989]{nelson1989categorical}
Nelson, D.~A. and Marler, P. (1989).
\newblock Categorical perception of a natural stimulus continuum: birdsong.
\newblock {\em Science}, 244(4907):976--978.

\bibitem[Okazawa et~al., 2021]{okazawa2021thegeometry}
Okazawa, G., Hatch, C., Mancoo, A., Machens, C., and Kiani, R. (2021).
\newblock The geometry of the representation of decision variable and stimulus
  difficulty in the parietal cortex.
\newblock {\em bioRxiv preprint 2021.01.04.425244}.

\bibitem[Olah, 2015]{olah2015visualizing}
Olah, C. (2015).
\newblock Visualizing representations: Deep learning and human beings.
\newblock Available at
  \url{colah.github.io/posts/2015-01-Visualizing-Representations/}. Accessed
  May 2021.

\bibitem[Padgett and Cottrell, 1998]{padgett1998simple}
Padgett, C. and Cottrell, G.~W. (1998).
\newblock A simple neural network models categorical perception of facial
  expressions.
\newblock In {\em Proceedings of the twentieth annual cognitive science
  conference}, pages 806--807. Citeseer.

\bibitem[Park and Kwak, 2016]{park2016analysis}
Park, S. and Kwak, N. (2016).
\newblock Analysis on the dropout effect in convolutional neural networks.
\newblock In {\em Asian conference on computer vision}, pages 189--204.
  Springer.

\bibitem[Pedregosa et~al., 2011]{scikit-learn}
Pedregosa, F., Varoquaux, G., Gramfort, A., Michel, V., Thirion, B., Grisel,
  O., Blondel, M., Prettenhofer, P., Weiss, R., Dubourg, V., Vanderplas, J.,
  Passos, A., Cournapeau, D., Brucher, M., Perrot, M., and Duchesnay, E.
  (2011).
\newblock Scikit-learn: Machine learning in {P}ython.
\newblock {\em Journal of Machine Learning Research}, 12:2825--2830.

\bibitem[Rennie et~al., 2014]{rennie2014annealed}
Rennie, S.~J., Goel, V., and Thomas, S. (2014).
\newblock Annealed dropout training of deep networks.
\newblock In {\em 2014 IEEE Spoken Language Technology Workshop (SLT)}, pages
  159--164. IEEE.

\bibitem[Repp, 1984]{Repp_1984}
Repp, B. (1984).
\newblock Categorical perception: issues, methods, findings.
\newblock In {\em Speech and Language: Advances in Basic Research and
  Practice}.

\bibitem[Russakovsky et~al., 2015]{ILSVRC15}
Russakovsky, O., Deng, J., Su, H., Krause, J., Satheesh, S., Ma, S., Huang, Z.,
  Karpathy, A., Khosla, A., Bernstein, M., Berg, A.~C., and Fei-Fei, L. (2015).
\newblock {ImageNet Large Scale Visual Recognition Challenge}.
\newblock {\em International Journal of Computer Vision (IJCV)},
  115(3):211--252.

\bibitem[Schmidhuber, 2015]{schmidhuber2015deep}
Schmidhuber, J. (2015).
\newblock Deep learning in neural networks: An overview.
\newblock {\em Neural networks}, 61:85--117.

\bibitem[Schwartz-Ziv and Tishby, 2017]{schwartz-ziv2017opening}
Schwartz-Ziv, R. and Tishby, N. (2017).
\newblock Opening the black box of deep neural networks via information.
\newblock {\em arXiv:1703.00810}.

\bibitem[Seung and Sompolinsky, 1993]{seung1993simple}
Seung, H.~S. and Sompolinsky, H. (1993).
\newblock Simple models for reading neuronal population codes.
\newblock {\em Proceedings of the national academy of sciences},
  90(22):10749--10753.

\bibitem[Seung et~al., 1992]{seung1992statistical}
Seung, H.~S., Sompolinsky, H., and Tishby, N. (1992).
\newblock Statistical mechanics of learning from examples.
\newblock {\em Physical Review A}, 45(8):6056--6091.

\bibitem[Simonyan and Zisserman, 2014]{simonyan2014very}
Simonyan, K. and Zisserman, A. (2014).
\newblock Very deep convolutional networks for large-scale image recognition.
\newblock {\em arXiv preprint arXiv:1409.1556}.

\bibitem[Spilsbury and Camps, 2019]{spilsbury2019don}
Spilsbury, T. and Camps, P. (2019).
\newblock Don't ignore dropout in fully convolutional networks.
\newblock {\em arXiv preprint arXiv:1908.09162}.

\bibitem[Srivastava et~al., 2014]{srivastava2014dropout}
Srivastava, N., Hinton, G., Krizhevsky, A., Sutskever, I., and Salakhutdinov,
  R. (2014).
\newblock Dropout: a simple way to prevent neural networks from overfitting.
\newblock {\em The journal of machine learning research}, 15(1):1929--1958.

\bibitem[Tijsseling and Harnad, 1997]{tijsseling1997warping}
Tijsseling, A. and Harnad, S. (1997).
\newblock Warping similarity space in category learning by backprop nets.
\newblock In {\em Proceedings of SimCat 1997: Interdisciplinary workshop on
  similarity and categorization}, pages 263--269. Department of Artificial
  Intelligence, Edinburgh University.

\bibitem[Tong et~al., 2008]{tong2008fusiform}
Tong, M.~H., Joyce, C.~A., and Cottrell, G.~W. (2008).
\newblock Why is the fusiform face area recruited for novel categories of
  expertise? a neurocomputational investigation.
\newblock {\em Brain Research}, 1202:14--24.

\bibitem[{Virtanen} et~al., 2020]{scipy}
{Virtanen}, P., {Gommers}, R., {Oliphant}, T.~E., {Haberland}, M., {Reddy}, T.,
  {Cournapeau}, D., {Burovski}, E., {Peterson}, P., {Weckesser}, W., {Bright},
  J., {van der Walt}, S.~J., {Brett}, M., {Wilson}, J., {Jarrod Millman}, K.,
  {Mayorov}, N., {Nelson}, A. R.~J., {Jones}, E., {Kern}, R., {Larson}, E.,
  {Carey}, C., {Polat}, {\.I}., {Feng}, Y., {Moore}, E.~W., {Vand erPlas}, J.,
  {Laxalde}, D., {Perktold}, J., {Cimrman}, R., {Henriksen}, I., {Quintero},
  E.~A., {Harris}, C.~R., {Archibald}, A.~M., {Ribeiro}, A.~H., {Pedregosa},
  F., {van Mulbregt}, P., and {Contributors}, S. .~. (2020).
\newblock {SciPy 1.0: Fundamental Algorithms for Scientific Computing in
  Python}.
\newblock {\em Nature Methods}, 17:261--272.

\bibitem[Wood, 1976]{wood1976discriminability}
Wood, C.~C. (1976).
\newblock Discriminability, response bias, and phoneme categories in
  discrimination of voice onset time.
\newblock {\em The Journal of the Acoustical Society of America},
  60(6):1381--1389.

\bibitem[Xin et~al., 2019]{xin2019sensory}
Xin, Y., Zhong, L., Zhang, Y., Zhou, T., Pan, J., and Xu, i.-l. (2019).
\newblock Sensory-to-category transformation via dynamic reorganization of
  ensemble tructures in mouse auditory cortex.
\newblock {\em Neuron}, 103(5):909--921.

\bibitem[Zhao et~al., 2019]{zhao2019equivalence}
Zhao, D., Yu, G., Xu, P., and Luo, M. (2019).
\newblock Equivalence between dropout and data augmentation: A mathematical
  check.
\newblock {\em Neural Networks}, 115:82--89.

\end{thebibliography}

\end{document}